\documentclass{article}

\usepackage{microtype}
\usepackage{graphicx}
\usepackage{subcaption}
\usepackage{booktabs} 

\usepackage{newfloat}
\usepackage{listings}
\usepackage{amsmath}
\usepackage[table]{xcolor}
\usepackage{booktabs}
\usepackage[listings, breakable, skins, most]{tcolorbox}
\usepackage{lipsum}
\usepackage{xcolor}

\usepackage{hyperref}
\usepackage{multirow} 

\newcommand{\colorCell}[1]{%
  \pgfmathparse{int(#1 * 100)}%
  \edef\temp{\noexpand\cellcolor{orange!\pgfmathresult}}%
  \temp #1%
}

\usepackage[preprint]{icml2026}

\usepackage{amsmath}
\usepackage{amssymb}
\usepackage{mathtools}
\usepackage{amsthm}

\usepackage[capitalize,noabbrev]{cleveref}

\theoremstyle{plain}

\theoremstyle{definition}

\theoremstyle{remark}

\usepackage[textsize=tiny]{todonotes}

\icmltitlerunning{LLMs as Self-Judges for Human-Aligned Evaluation}

\begin{document}

\twocolumn[
  \icmltitle{LLMs Judge Themselves: A Game-Theoretic Framework \\ for Human-Aligned Evaluation}

  \icmlsetsymbol{equal}{*}

  \begin{icmlauthorlist}
    \icmlauthor{Yang Gao}{equal,bit,sait}
    \icmlauthor{Yuhang Liu}{equal,bit,sait}
    \icmlauthor{Siyu Miao}{bit}
    \icmlauthor{Xinyue Liang}{bit}
    \icmlauthor{Zhengyang Liu}{bit,sait}
    \icmlauthor{Heyan Huang}{bit,sait}
  \end{icmlauthorlist}

  \icmlaffiliation{bit}{
    Beijing Institute of Technology, Beijing, China
  }
  \icmlaffiliation{sait}{
    Southeast Academy of Information Technology, China
  }

  \icmlcorrespondingauthor{Zhengyang Liu}{zhengyang@bit.edu.cn}

  \icmlkeywords{Large Language Models, Evaluation, Game Theory}

  \vskip 0.3in
]



\printAffiliationsAndNotice{\icmlEqualContribution}

\begin{abstract}
Ideal or real—that is the question. In this work, we explore whether principles from game theory can be effectively applied to the evaluation of large language models (LLMs). This inquiry is motivated by the growing inadequacy of conventional evaluation practices, which often rely on fixed-format tasks with reference answers and struggle to capture the nuanced, subjective, and open-ended nature of modern LLM behavior. To address these challenges, we propose a novel alternative: automatic mutual evaluation, where LLMs assess each other's output through self-play and peer review. These peer assessments are then systematically compared with human voting behavior to evaluate their alignment with human judgment. Our framework incorporates game-theoretic voting algorithms to aggregate peer reviews, enabling a principled investigation into whether model-generated rankings reflect human preferences. Empirical results reveal both convergences and divergences between theoretical predictions and human evaluations, offering valuable insights into the promises and limitations of mutual evaluation. To the best of our knowledge, this is the first work to jointly integrate mutual evaluation, game-theoretic aggregation, and human-grounded validation for evaluating the capabilities of LLMs.
\end{abstract}


\section{Introduction}

\begin{figure}[tp]
    \centering
    \includegraphics[width=0.4\textwidth,height=4.6cm]{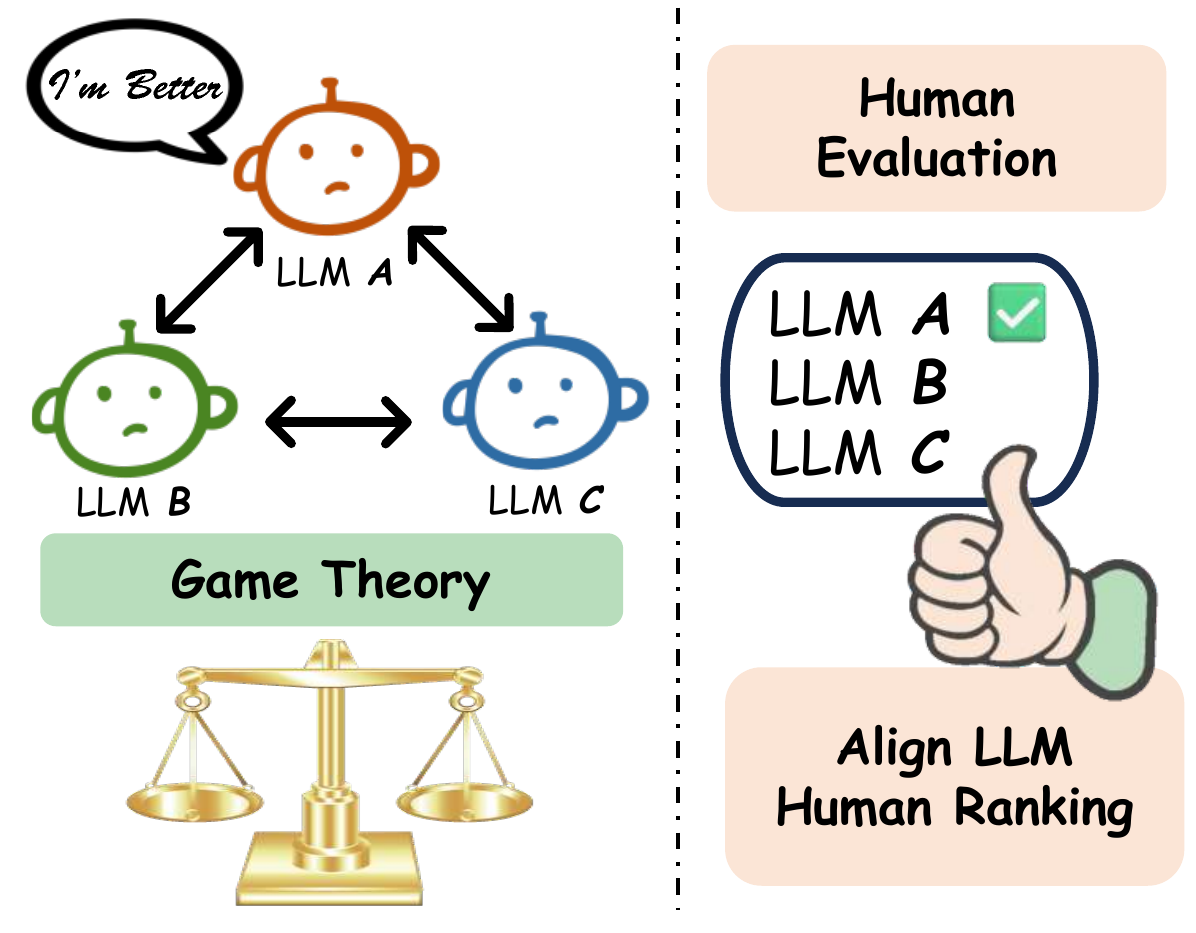}
    \caption{Illustration of game-theoretic peer evaluation for LLM performance ranking.}
    \label{fig:peer-eval-illustration}
\end{figure}

As Large Language Models (LLMs)~\cite{qiu2025how, DBLP:journals/corr/abs-2501-12948, DBLP:journals/corr/abs-2412-16720} are increasingly deployed in real-world applications, evaluation has shifted from isolated performance measurement to comparative model ranking~\cite{liu2025aligninghumanjudgementrole, DBLP:conf/icml/ChiangZ0ALLZ0JG24}. However, existing evaluation methodologies~\cite{lin2023llmevalunifiedmultidimensionalautomatic} often simplify assessment to single-model judgments or scalar metrics, resulting in misalignment with human comparative judgments.
Humans typically navigate such contexts by aggregating partial rankings from diverse perspectives, then synthesizing them into a globally optimized consensus~\cite{yuan2025casestudyscalablecontent, lyon2013wisdom}. In contrast, LLM-based evaluations tend to produce singular, context-insensitive outputs, lacking mechanisms for balancing competing preferences or integrating different judgments~\cite{Wang_2025}. To bridge this gap, this paper introduces a game-theoretic ranking framework into LLM evaluation, designed to emulate human-like consensus formation through preference aggregation and optimization. Our approach aims to enhance the alignment, fairness, and robustness of LLM assessments, especially in complex, subjective, or multi-stakeholder environments. 

\textbf{Game theory}, as a theoretical framework for analyzing strategy selection~\cite{hansen2025stochasticgameslimitedpublic} and utility distribution~\cite{DBLP:conf/aaai/MirfakharWZZH25} among interacting agents, emphasizes mechanisms of competition, cooperation, and dynamic feedback~\cite{DBLP:conf/icml/Zhu0ZX024, DBLP:journals/tacl/FabbriKMXSR21}. These properties naturally align with the inherent capability differences, judgment conflicts, and task adaptability among LLMs, enabling game-theoretic approaches to effectively simulate complex human evaluation processes~\cite{pasch2025aivshumanjudgment, DBLP:journals/corr/abs-2412-16720}. Recent surveys have noted this emerging connection between game-theoretic perspectives and LLMs~\cite{DBLP:conf/ijcai/SunWCC25}. Conversely, LLMs also offer an ideal testbed for empirically validating theoretical insights from game theory. Motivated by these observations, we ask: \textit{Can game-theoretic methodologies yield model rankings that align with human judgments in evaluating the capabilities of LLMs?} Figure~\ref{fig:peer-eval-illustration} provides an illustrative overview of this central question.

To validate the aforementioned hypothesis, one key challenge lies in the presence of self-preference bias~\cite{DBLP:conf/acl/LiuML24, dietz2025llmevaluationtropesperspectivesvalidity}, where models tend to favor their own outputs when serving as evaluators. Such bias can distort evaluation outcomes and undermine their fairness and reliability. To address this, we adopt an automated peer evaluation mechanism utilizing large models~\cite{DBLP:journals/corr/abs-2410-12265, DBLP:journals/corr/abs-2401-15641} to simulate a game-theoretic assessment process. 
Specifically, we formalize this process as a game-theoretic voting system and introduce the \textit{decentralized peer ranking} framework, where each LLM simultaneously serves as both an evaluatee and an evaluator without relying on a central judge. For each evaluation problem, evaluators produce complete rankings over candidate model outputs, forming a ranking matrix that summarizes judgments across evaluators. Based on this ranking matrix, we apply a suite of aggregation algorithms~\cite{DBLP:conf/naacl/ZhangWLCLXHZPLOZ25,DBLP:journals/corr/abs-2411-03390,DBLP:journals/jacm/CharikarRWW24,DBLP:conf/wine/XiaZ22,DBLP:conf/opodis/WoodMP24a} to derive a \textit{consensus ranking} that reflects the aggregated preferences of all evaluators, and systematically compare aggregation methods in terms of their alignment with human evaluations. 


This decentralized peer ranking framework replaces centralized judgment with mutual model-based assessment, enabling collective evaluation without a single authoritative judge, while its reliability and fairness depend on several key factors. Before presenting the experimental design, it is important to clarify the core research objectives our study seeks to address.

\textbf{During the peer evaluation, how well do different game-theoretic peer evaluation algorithms align with human judgments in ranking LLMs?} 
We find that aggregation-based ranking methods, which derive a consensus ranking from multiple evaluator-provided rankings, achieve substantially stronger and more stable alignment with human judgments than single-model evaluations. Among these methods, consensus-based aggregation method, particularly the Kemeny--Young method~\cite{kemeny1959mathematics}, consistently exhibit the highest alignment across benchmarks. Moreover, such game-theoretic aggregation approaches also outperform rankings induced by standard accuracy-based benchmarks, highlighting their advantage in capturing human-aligned preferences beyond correctness alone. Importantly, this scalability is achieved without changing the aggregation method, indicating that game-theoretic ranking remains effective even when only sparse comparisons are available.

\textbf{Can game-theoretic ranking algorithms mitigate such self-bias and produce more reliable evaluation rankings?} 
Consistent with prior studies~\cite{DBLP:conf/acl/LiuML24, dietz2025llmevaluationtropesperspectivesvalidity}, our results confirm that LLMs exhibit systematic self-preference bias, with self-assigned rankings showing weak and unstable alignment with human judgment. We further find that game-theoretic aggregation substantially reduces the impact of this bias. In particular, consensus-based aggregation produces stable rankings that remain well aligned with human preferences even when self-assigned rankings are included, demonstrating the robustness of game-theoretic methods to self-preference bias.

\textbf{Are there specific LLM capabilities where game-theoretic evaluation exhibits varying alignment with human judgment?} LLMs possess diverse capabilities across a wide range of tasks, from logical reasoning to creative generation. 
Our results show that alignment with human judgment varies substantially across different capability domains. Game-theoretic aggregation achieves the strongest and most stable alignment on tasks with clear and objective correctness signals, such as mathematical reasoning tasks, while exhibiting weaker and more variable alignment on more subjective tasks, including creative writing and linguistically nuanced benchmarks. Nevertheless, even in such settings, aggregate rankings can still capture human consensus at the dataset level overall.



In summary, we propose a decentralized, game-theoretic peer ranking framework in which each LLM simultaneously serves as an evaluatee and an evaluator, with each evaluator producing a ranking over candidate models and a consensus ranking obtained by aggregating these individual rankings using principled aggregation algorithms. Using this framework, we conduct a systematic analysis of aggregation-based ranking methods across multiple benchmarks and evaluation settings, comparing their alignment with human judgments, robustness to self-preference bias, and consistency across capability domains. Our results show that consensus-based aggregation methods, particularly the Kemeny--Young method, consistently produce rankings that are more stable and more closely aligned with human preferences than both single-model evaluations and accuracy-based benchmarks. We further demonstrate that game-theoretic aggregation substantially reduces the impact of self-preference bias and remains reliable even when self-assigned rankings are included, while revealing clear differences in reliability across task types, with the strongest performance on tasks with clear and objective correctness signals. Taken together, these findings highlight the importance of treating LLM evaluation as a structured aggregation and decision-making problem, and suggest that future evaluation protocols should prioritize consensus-oriented and bias-aware aggregation mechanisms to achieve fair and human-aligned assessment.

\begin{figure*}[htp]
    \centering
    \includegraphics[width=0.8\textwidth, height=8cm]{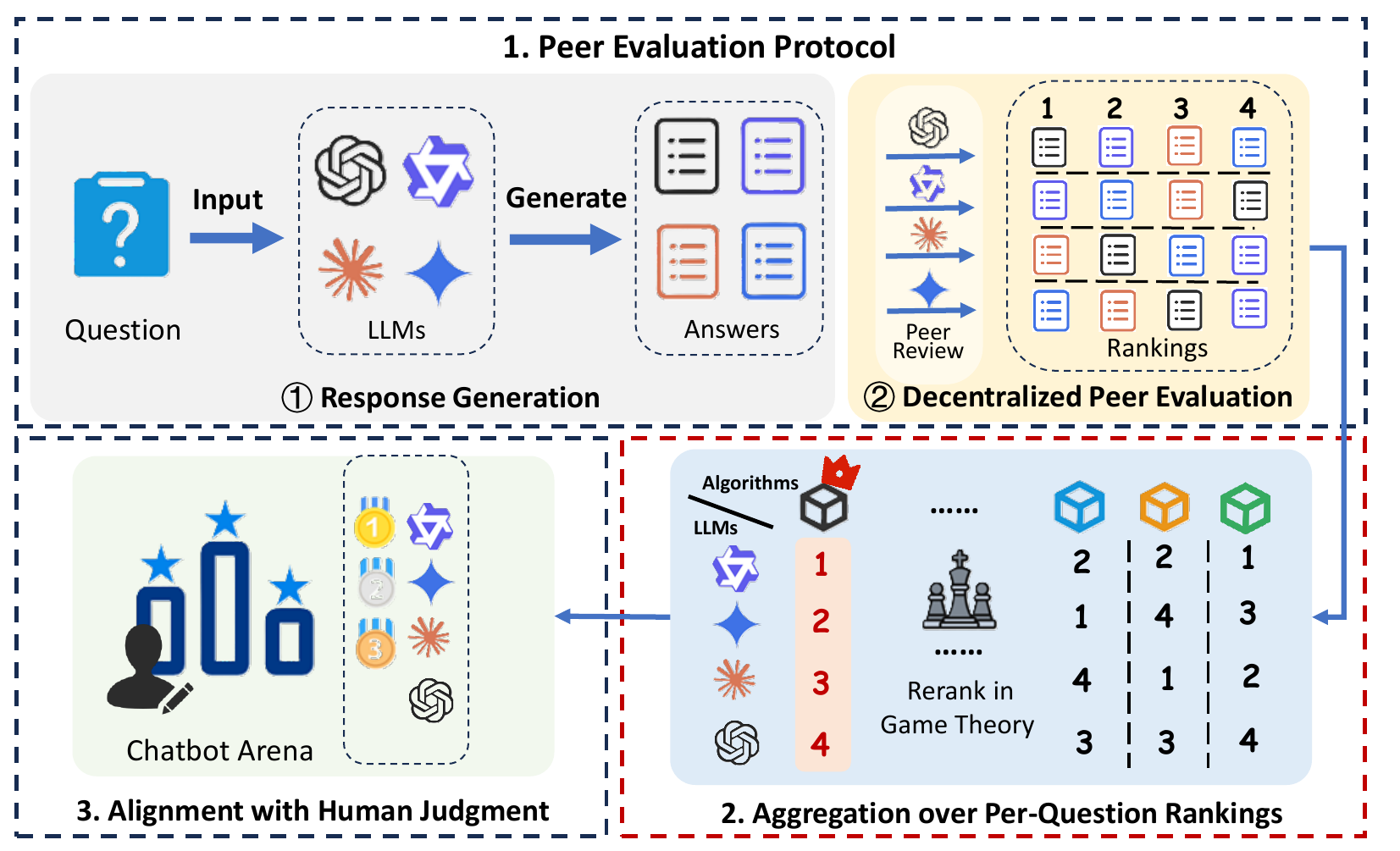}
    \caption{The proposed framework for game-theoretic evaluation of LLMs.}
    \label{fig:mutul-eval}
\end{figure*}

\section{Related Work}

\subsection{Evaluation Methods for LLMs}
Evaluating LLMs remains challenging due to their rapidly expanding capabilities. Existing approaches can be broadly divided into two paradigms: human annotated evaluation and LLM as a judge evaluation. Traditional benchmarks such as MMLU~\cite{DBLP:conf/iclr/HendrycksBBZMSS21} and GSM8K~\cite{DBLP:conf/icml/ChiangZ0ALLZ0JG24} rely on curated datasets with reference answers and human or rule based grading, offering consistency and reproducibility for model comparison. However, increasing benchmark saturation has raised concerns about the ability of static test sets to reflect generalizable model performance~\cite{DBLP:journals/corr/abs-2510-26852, DBLP:journals/corr/abs-2510-26768}. An alternative paradigm uses LLMs as evaluators, enabling scalable assessment for open ended and subjective tasks. Prior work often selects a single LLM as the judge~\cite{DBLP:conf/cikm/ChuATL024}, which may introduce evaluator bias and limit perspective. In contrast, our approach adopts a decentralized evaluation setting in which all participating models contribute evaluations, and a game theoretic aggregation method is applied to derive a consensus ranking, thereby reducing individual bias and yielding a more robust assessment.

\subsection{Game-Theoretic Approaches to Evaluation}
Game theoretic methods provide a general evaluation framework~\cite{DBLP:journals/corr/abs-2402-12348} based on relative comparisons, without relying on fixed ground truth labels. Prior work in this area largely follows two directions. The first constructs game inspired environments, such as matrix games or auctions, to study decision making and strategic behavior of language models under different payoff structures~\cite{DBLP:journals/corr/abs-2508-03368}. While effective for behavioral analysis, these settings are mainly used for probing model behavior rather than performance evaluation or ranking. The second direction draws on preference aggregation techniques from social choice theory to derive consensus rankings from pairwise comparisons, which improve robustness and reduce evaluator bias in subjective settings. For example, \citet{DBLP:conf/naacl/ZhangWLCLXHZPLOZ25} apply an enhanced Borda count to re rank mathematical reasoning responses. Similar aggregation strategies have also been adopted in LLM evaluation and related domains where absolute ground truth is unavailable.

\section{Methodology}
\label{sec:method}
\subsection{Overview of Our Evaluation Framework}

Conventional evaluation methods for LLMs often rely on comparisons against predefined reference answers, which emphasize correctness with respect to fixed outputs. While effective for many benchmark settings, such approaches do not explicitly model the comparative and preference-based nature of human judgment that arises in many real-world evaluation scenarios.


As shown in Figure~\ref{fig:mutul-eval}, we propose an automated peer evaluation framework in which LLMs assess the outputs of all candidate models, including their own. Each model acts as both an evaluatee and an evaluator, producing rankings over peer-generated responses. These decentralized judgments are aggregated using game theoretic voting algorithms to yield a consensus ranking over all candidate models at each question, which supports analysis of alignment with human preferences. The framework consists of three core components: the peer evaluation protocol, aggregation of per question rankings, and alignment with human judgment.

\subsection{Peer Evaluation Protocol}
\label{sec:peerreview}

Let \( \mathcal{M} = \{LLM_1, \ldots, LLM_M\} \) denote a collection of \( M \) language models, and let \( \mathcal{Q} = \{q_1, \ldots, q_N\} \) be a shared set of evaluation questions. For each model \( LLM_i \in \mathcal{M} \), we collect a set of responses
\( A_i^{\mathcal{Q}} = \{a_{ij}\}_{j=1}^N \),
where \( a_{ij} \) denotes the response produced by \( LLM_i \) for question \( q_j \).

As illustrated in the upper portion of Figure~\ref{fig:mutul-eval}, the evaluation procedure consists of two stages.

\textbf{Response Generation.}
Each model independently responds every question \( q_j \in \mathcal{Q} \), producing a response \( a_{ij} \) for each \( LLM_i \in \mathcal{M} \).

\textbf{Decentralized Peer Evaluation.}
For each question \( q_j \), every model \( LLM_i \) simultaneously serves as an evaluator and is presented with a randomly ordered and anonymized set of responses \( A^{(j)} = \{a_{1j}, a_{2j}, \dots, a_{Mj}\} \), including its own. The evaluator provides a complete ranking \( \pi_i^{(j)} \in \mathcal{S}_M \) over the candidate responses based on their quality with respect to the given question, where \( \mathcal{S}_M \) denotes the set of all permutations over the \( M \) candidate models. The ranking \( \pi_i^{(j)} \) represents the model’s subjective preference ordering for question \( q_j \), analogous in form to a human-provided ranking. Collectively, the set \( \Pi^{(j)} = \{\pi_i^{(j)}\}_{i=1}^M \) constitutes a decentralized preference profile for question \( q_j \).


\subsection{Aggregation over Per-Question Rankings}
\label{sec:aggregation}
Given the decentralized preference profiles \( \Pi^{(j)} \) obtained from peer evaluation for a given evaluation question \( q_j \), the aggregation problem maps the preference profile \( \Pi^{(j)} \) to a consensus ranking \(\pi^{(j)} \in \mathcal{S}_M\), which reconciles potentially conflicting judgments from different evaluators.

Since rankings may vary across evaluators due to differences in model capabilities, evaluation noise, or systematic biases, the aggregation method must be robust and consistent. We consider multiple ranking aggregation algorithms, including Borda count, Copeland voting, Kemeny--Young and so on. In our experiments, aggregation-based methods consistently exhibit stronger alignment with human rankings than individual model judgments. Among them, the Kemeny--Young method achieves the highest and most stable alignment with human preferences and is therefore adopted as the default aggregation strategy. For each question \( q_j \), Kemeny--Young identifies the ranking \( \pi^{(j)} \) that minimizes the total Kendall--Tau distance to the input rankings in \( \Pi^{(j)} \), equivalently minimizing total pairwise disagreement. Alternative aggregation strategies are discussed in Appendix~\ref{appendix:rankers}.

\subsection{Alignment with Human Judgment}
\label{sec:alignment}
To evaluate the validity of our framework, we compare the rankings produced by game-theoretic evaluation method against human preference ranking. Specifically, we align our rankings with those derived from Chatbot Arena \citep{DBLP:conf/icml/ChiangZ0ALLZ0JG24}, a large-scale crowdsourced evaluation platform in which human annotators express preferences through pairwise comparisons of model outputs. Aggregating these pairwise votes across annotators and prompts yields a consensus ranking that reflects collective human preferences over LLMs. This aggregated ranking has been widely adopted as a reference for human-aligned model comparison and therefore serves as our benchmark for alignment analysis. We compute several alignment metrics to quantify the similarity between our game-theoretic rankings and those derived from Chatbot Arena. Specifically, we consider the \textbf{Pearson correlation coefficient}, which measures the linear correlation between model scores under the two rankings, and \textbf{Kendall's \( \tau \)}~\cite{kendall1948rank}, which measures the rank correlation between the two rankings.

High alignment scores indicate that decentralized model-driven evaluation can effectively approximate aggregated human judgments, offering a scalable and label-free alternative to traditional human evaluation pipelines.

\begin{table*}[htbp]
\centering
\caption{Distribution of \textbf{Pearson} and \textbf{Kendall} correlation coefficients between automated rankings and human ranking on the \texttt{GenOverall} dataset. Correlations are computed at the micro level as the alignment between model rankings and human rankings for each question. The left columns show the results for individual LLMs, while the right columns report outcomes after applying different aggregation algorithms. Summary statistics include the mean, standard deviation, minimum, quartiles, and maximum. Higher values indicate stronger alignment with human judgment. Color intensity reflects correlation strength. Abbreviations: Avg(average voting), Dod(Dodgson), Cop(Copeland), Bor(Borda), Irv(instant-runoff voting), Spm(Spearman), Kem(Kemeny-Young), Ken(Kendall). Models: 4o11(gpt-4o-20241120), 4o05(gpt-4o-20240513), 4o08(gpt-4o-20240806), Sn10(claude-3.5-Sonnet-20241022), Hk10(claude-3.5-haiku-20241022), Op02(claude-3-opus-20240229).
}
\resizebox{\textwidth}{!}{
\begin{tabular}{ll|cccccc|cccccccc}
\toprule
& & \textbf{4o11} & \textbf{Sn10} & \textbf{4o05} & \textbf{4o08} & \textbf{Op02} & \textbf{Hk10} & \textbf{Avg} & \textbf{Dod} & \textbf{Cop} & \textbf{Bor} & \textbf{Irv} & \textbf{Spm} & \textbf{Kem} & \textbf{Ken} \\
\midrule

\multirow{7}{*}{\rotatebox{90}{\texttt{Pearson}}} & \textbf{Mean} & 0.296          & 0.156          & 0.311          & 0.313          & 0.169          & 0.024           & 0.300          & 0.313          & 0.342          & 0.302          & -0.151         & 0.289          & 0.367          & 0.345          \\
                  & \textbf{Std.} & 0.494          & 0.511          & 0.511          & 0.499          & 0.504          & 0.493           & 0.530          & 0.521          & 0.536          & 0.526          & 0.475          & 0.517          & 0.540          & 0.539          \\
                  & \textbf{Min}  & -0.829         & -0.829         & -0.771         & -0.771         & -0.771         & -0.714          & -0.784         & -0.771         & -0.771         & -0.771         & -1.000         & -0.771         & -0.771         & -0.771         \\
                  & \textbf{25\%} & \colorCell{0.014} & -0.271         & -0.043         & \colorCell{0.029} & -0.314         & -0.257          & -0.203         & -0.214         & -0.214         & -0.214         & -0.657         & -0.214         & -0.214         & -0.214         \\
                  & \textbf{50\%} & \colorCell{0.343} & \colorCell{0.257} & \colorCell{0.371} & \colorCell{0.343} & \colorCell{0.343} & {-0.086} & \colorCell{0.504} & \colorCell{0.543} & \colorCell{0.514} & \colorCell{0.514} & -0.143         & \colorCell{0.457} & \colorCell{0.571} & \colorCell{0.486} \\
                  & \textbf{75\%} & \colorCell{0.671} & \colorCell{0.486} & \colorCell{0.671} & \colorCell{0.671} & \colorCell{0.500} & \colorCell{0.371}  & \colorCell{0.683} & \colorCell{0.714} & \colorCell{0.771} & \colorCell{0.671} & \colorCell{0.200} & \colorCell{0.671} & \colorCell{0.786} & \colorCell{0.771} \\
                  & \textbf{Max}  & 1.000          & 0.943          & 1.000          & 1.000          & 0.943          & 0.943           & 0.968          & 1.000          & 1.000          & 1.000          & 0.714          & 0.943          & 1.000          & 1.000  \\
\hline
\multirow{7}{*}{\rotatebox{90}{\texttt{Kendall}}} & \textbf{Mean} & 0.245     & 0.131     & 0.291     & 0.261     & 0.139     & 0.092     & 0.322     & 0.370     & 0.322     & 0.348     & -0.237    & 0.328     & 0.362     & 0.350     \\
   & \textbf{Std.} & 0.336     & 0.326     & 0.343     & 0.357     & 0.317     & 0.345     & 0.359     & 0.340     & 0.328     & 0.358     & 0.386     & 0.368     & 0.355     & 0.353     \\
   & \textbf{Min}  & -0.600    & -0.600    & -0.467    & -0.467    & -0.467    & -0.733    & -0.467    & -0.467    & -0.467    & -0.467    & -0.867    & -0.467    & -0.467    & -0.467    \\
   & \textbf{25\%} & \colorCell{0.067} & -0.067    & \colorCell{0.067} & \colorCell{0.067} & -0.067    & -0.133    & \colorCell{0.067} & \colorCell{0.133} & \colorCell{0.133} & \colorCell{0.067} & -0.467    & \colorCell{0.067} & \colorCell{0.133} & \colorCell{0.067} \\
   & \textbf{50\%} & \colorCell{0.333} & \colorCell{0.133} & \colorCell{0.333} & \colorCell{0.200} & \colorCell{0.200} & \colorCell{0.067} & \colorCell{0.414} & \colorCell{0.467} & \colorCell{0.467} & \colorCell{0.467} & -0.333    & \colorCell{0.333} & \colorCell{0.467} & \colorCell{0.467} \\
   & \textbf{75\%} & \colorCell{0.467} & \colorCell{0.333} & \colorCell{0.467} & \colorCell{0.467} & \colorCell{0.433} & \colorCell{0.333} & \colorCell{0.552} & \colorCell{0.600} & \colorCell{0.600} & \colorCell{0.600} & \colorCell{0.067} & \colorCell{0.600} & \colorCell{0.600} & \colorCell{0.600} \\
   & \textbf{Max}  & 0.867     & 0.733     & 1.000     & 1.000     & 0.733     & 0.867     & 1.000     & 1.000     & 0.867     & 1.000     & 0.600     & 1.000     & 1.000     & 1.000   \\
\bottomrule
\end{tabular}
}

\label{tab:main-genoverall}
\end{table*}

\section{Experiments and Analysis}

\subsection{Experimental Setups}

\paragraph{Datasets.} 
We evaluate model reasoning across a diverse set of benchmarks, including GSM8K~\cite{DBLP:conf/icml/ChiangZ0ALLZ0JG24}, MMLU~\cite{DBLP:conf/iclr/HendrycksBBZMSS21}, GPQA~\cite{DBLP:journals/corr/abs-2311-12022}, CEval~\cite{DBLP:conf/nips/HuangBZZZSLLZLF23}, IFEval~\cite{DBLP:journals/corr/abs-2311-07911}, MBPP~\cite{DBLP:journals/corr/abs-2108-07732}, and the Creative Writing benchmark~\cite{DBLP:journals/corr/abs-2503-05244}. All datasets are subsampled for tractability while preserving diversity across domains and reasoning types. To further mitigate the effects of benchmark saturation, we additionally prompt \texttt{GPT-4} to construct 3 synthetic evaluation sets, namely GenOverall, GenMath, and GenChinese, which are designed to cover broad reasoning abilities, mathematical reasoning, and Chinese language understanding, respectively. Details of the dataset construction process are provided in Appendix~\ref{appendix:datasets}.

\paragraph{Models.}
Our experiments include 6 recent LLMs from OpenAI (GPT-4o variants) and Anthropic (Claude-3 and 3.5 series), chosen for their consistently top-tier performance in recent benchmarks.

\paragraph{Human Preference Reference.}
To evaluate how well our framework aligns with human preferences, we compare the resulting model rankings against those from Chatbot Arena~\cite{DBLP:conf/icml/ChiangZ0ALLZ0JG24}. Since Chatbot Arena provides system-level rankings across multiple domains, we match each benchmark to the most relevant leaderboard subset. Specifically, we use the GSM8K and GenMath for \textit{math-specific} ranking, the MMLU, GPQA and GenOverall for \textit{overall} ranking, the CEval and GenChinese for \textit{Chinese} ranking, the IFEval for \textit{instruction-following} ranking, and the MBPP for \textit{code-related} rankings. This task-to-domain mapping ensures that our comparisons reflect the most appropriate human preference signals. Alignment is quantified using both Pearson and Kendall~\cite{kendall1948rank} correlation coefficients between our aggregated rankings and those from Chatbot Arena.

\paragraph{Evaluation Levels.}
To assess how well model-generated rankings align with human preferences, we evaluate correlation at two complementary levels: \textit{micro-level} and \textit{macro-level} respectively.

\textbf{Micro-level correlation} measures alignment at the level of individual questions. For each question, we compute the correlation between rankings produced by a given aggregation method and the corresponding human reference, yielding a set of correlation scores. We then analyze their distribution using summary statistics such as the mean, standard deviation, and percentiles to assess both alignment quality and consistency. For example, a higher 25th percentile generally indicates that most questions exhibit strong alignment with human judgments.
    
\textbf{Macro-level correlation} computes the average of per-question rankings to obtain a single overall ranking of models for the entire dataset, and then computes a single correlation score with the human-provided global ranking. This captures overall alignment at the dataset level but does not reflect per-question variability.


\paragraph{Ranking Algorithms.} To produce a final ranking over models from decentralized peer evaluations, we implement nine classical aggregation methods from social choice theory and voting literature: Average, Borda Count~\cite{saari1985optimal}, Copeland~\cite{saari1996copeland}, Dodgson~\cite{brandt2009some}, Instant-Runoff Voting (IRV)~\cite{brandt2009some}, Kemeny-Young~\cite{kemeny1959mathematics}, Kendall~\cite{kendall1948rank}, and Spearman. These algorithms take as input either ranking lists or pairwise preferences collected across evaluation questions and produce an overall consensus ranking of models. They cover a broad range of principles, including positional scoring methods such as Borda, pairwise comparison approaches such as Copeland, and consistency-based objectives such as Kemeny. Further implementation details are provided in Appendix~\ref{appendix:rankers}.

\subsection{Q1: Can Game-Theoretic Evaluation Align with Human Judgment?}
\label{exp:Q1}
To evaluate alignment with human judgment, we measure the alignment between rankings produced by different aggregation algorithms and the Chatbot Arena leaderboard~\cite{DBLP:conf/icml/ChiangZ0ALLZ0JG24} using Pearson and Kendall correlations. To reduce the impact of benchmark saturation, we primarily report results on \texttt{GenOverall} (Table~\ref{tab:main-genoverall}), while results on additional datasets are provided in Appendix~\ref{appendix:appendxq1}.

\begin{table}[tbp]
\centering
\caption{Alignment Between \textbf{Macro-Level} Model Rankings and Human Preferences. \textbf{Acc} denotes the \textbf{correlation score} between human rankings and the model rankings induced by overall accuracy. \textbf{Kem} denotes the \textbf{correlation score} between human rankings and the model rankings derived from Kemeny-young aggregation over per-question peer evaluations. Pearson (Prs) and Kendall (Ken) coefficients are reported, with higher values indicating stronger alignment with human judgments from Chatbot Arena.}
\resizebox{0.47\textwidth}{!}{
\begin{tabular}{lccccccc}
\toprule
 &              & CEval  & Writing & GSM8K & GPQA & MMLU \\
\hline  
\multirow{2}{*}{Prs}   & \textbf{Acc} & 0.227 & -                & 0.319          & 0.311         & 0.932         \\
& \textbf{Kem} & 0.714 & 0.914  & 0.941  & 0.907    & 0.919         \\
\hline
\multirow{2}{*}{Ken} & \textbf{Acc} & 0.138          &                 -                & 0.086          & 0.298         & 0.894    \\
 & \textbf{Kem} & 0.467 & 0.867            & 0.867          & 0.733         & 0.733         \\
\bottomrule
\end{tabular}}

\label{table:macro}
\end{table}

\paragraph{Re-ranking Methods Achieve Higher Alignment with Human Judgment than Single-Model Evaluations.}
As shown in Table~\ref{tab:main-genoverall}, most rank aggregation algorithms achieve substantially stronger alignment with human judgment than individual models, with the exception of IRV. This advantage is evident in both higher average correlations and greater stability. For example, the \texttt{Kemeny} method achieve a median Pearson correlation of 0.571, substantially outperforming the strongest individual model, \texttt{gpt-4o-20241120}, which attains a median of 0.343 with considerably higher variance. These results indicate that aggregation methods not only improve average alignment with human preferences, but also yield more consistent evaluation outcomes across questions.

\paragraph{Kemeny Aggregation Emerges as the Most Effective Re-ranking Strategy.}
On the aggregation side, the \texttt{Kemeny-Young} algorithm exhibits strong performance, achieving the highest median correlations with human preferences and relatively narrow interquartile ranges, indicating both high alignment and low variance. In contrast, methods such as \texttt{IRV} exhibit lower mean correlations and wider distribution spreads, suggesting greater inconsistency in capturing human-aligned rankings. Across Tables~\ref{tab:q1-gsm8k}-\ref{tab:q4-math}, \texttt{Kendall} and \texttt{Kemeny--Young} exhibit similar performance in most settings. However, \texttt{Kemeny--Young} shows more consistent generalization across benchmarks, making it the most robust aggregation method overall.

\paragraph{Kemeny Aggregation Achieves Stronger Alignment with Human Judgment than Accuracy-Based Rankings.}
While overall accuracy is commonly used to summarize a model’s performance, its alignment with human preferences remains unclear. To address this, we evaluate \textbf{macro-level correlation}, which measures the correlation between model rankings and human judgments at the dataset level. We compare rankings based on overall accuracy with those obtained through game theoretic aggregation of per question peer evaluations. As shown in Table~\ref{table:macro}, per question rankings are first derived using the \texttt{Kemeny-Young} algorithm and then averaged to produce a macro level ranking, which is compared against human preferences. Across all benchmarks, this approach consistently achieves higher alignment with human judgments than accuracy based rankings. Notably, on the \texttt{Creative Writing} dataset, where no definitive reference answers are available, game theoretic aggregation continues to show strong correlation with human assessments, highlighting its robustness in subjective evaluation settings.

Among individual models, performance varies across datasets: \texttt{gpt-4o-20240513} performs best on \textsc{GenOverall} (Table~\ref{tab:main-genoverall}), while \texttt{gpt-4o-20241120} leads on the remaining datasets, as shown in Tables~\ref{tab:q1-gsm8k}--\ref{tab:q4-math}. In contrast, models such as \texttt{claude-3.5-sonnet} and \texttt{claude-3.5-haiku} exhibit lower correlations and higher variability, indicating less consistent alignment with human rankings. Overall, these results further highlight the benefit of game-theoretic rank aggregation, which not only improves alignment with human preferences relative to individual models but also yields more consistent and robust rankings across evaluation settings.

\begin{figure}[t]
    \centering
    \includegraphics[width=0.95\linewidth]{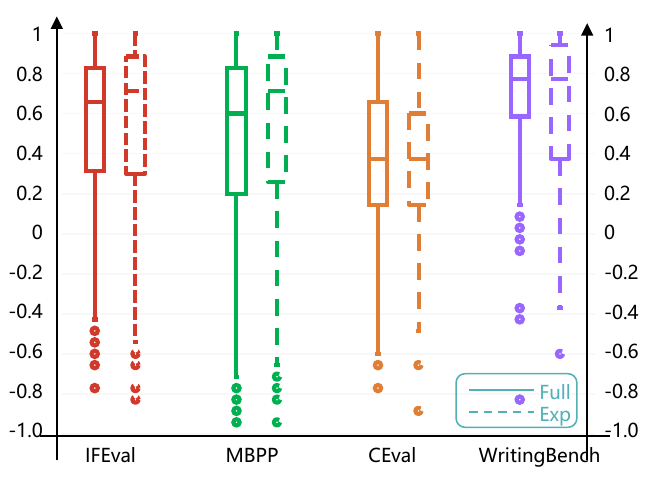}
    \caption{
    Pearson correlation distributions between human rankings and
    Kemeny--Young aggregated rankings under the sparse expander-graph setting.
    Solid boxes correspond to the full comparison graph, while dashed boxes
    correspond to the expander-based sparse graph. 
    }
    \label{fig:expander_scalability}
\end{figure}

\paragraph{Scalable Evaluation with Expander Graphs.}
To improve scalability, we adopt an extension based on expander graphs~\cite{deac2022expander}, which provide sparse yet well-connected comparison structures. Specifically, we consider two evaluation settings that differ in the density of model comparisons. In the \textbf{Full} setting, each evaluator produces a complete ranking over all $M$ candidate LLMs, resulting in a dense comparison graph with $O(M^2)$ total comparisons. In contrast, in the \textbf{Exp} setting, each evaluator is only required to rank a small subset of $K$ models (we use $K=3$ in all experiments), sampled according to an expander graph construction. This yields a sparse yet well-connected comparison graph with total complexity reduced to $O(KM)$, where $K \ll M$. Figure~\ref{fig:expander_scalability} compares the Pearson correlation between human rankings and Kemeny--Young aggregated rankings under the Full and Exp settings. Despite the drastic reduction in per-evaluator comparison load, from ranking 6 candidate models to ranking only 3, the resulting aggregated rankings remain highly consistent with those obtained under the full comparison graph. Across all benchmarks, the correlation distributions under the expander-based sparse setting closely match those of the full setting, indicating that substantial sparsification of evaluator judgments does not degrade alignment with human judgment. These results demonstrate that the proposed peer-ranking framework can scale to large model pools while preserving ranking stability and human alignment. Additional experimental details are provided in Appendix~\ref{append:full}.

\begin{table*}[htbp]
\centering
\caption{\textbf{Model Rankings} under Different Evaluation Protocols. This table reports model rankings under four evaluation protocols (SR, PR, SIA, and SFA) on the GenOL (GenOverall), GenMT (GenMath) and GenCN(GenChinese), where lower values indicate better performance (1 = best, 6 = worst). }
\renewcommand\arraystretch{1.1}
\resizebox{\textwidth}{!}{
\begin{tabular}{cccccccc}
\toprule
Dataset                & Method & 4o-1120 & Sn-1022 & 4o-0513 & 4o-0806 & Hk-1022 & Op-0229 \\
\hline
\multirow{4}{*}{GenOL} & SR  & 2.521         & 2.313         & 3.625         & 4.146         & 3.021         & 3.208         \\
                  & PR  & 2.929(+0.408)\textuparrow & 2.433(+0.121)\textuparrow & 3.733(+0.108)\textuparrow & 4.342(+0.196)\textuparrow & 3.925(+0.904)\textuparrow & 4.071(+0.863)\textuparrow          \\ \cline{2-8}
 & SIA & 2.396         & 2.333         & 3.479         & 4.604         & 3.979         & 4.208         \\
                  & SFA & 2.708(+0.313)\textuparrow & 2.354(+0.021)\textuparrow & 3.604(+0.125)\textuparrow & 4.521(-0.083) & 4.021(+0.042)\textuparrow & 4.125(-0.083)         \\
\hline
\multirow{4}{*}{GenMT}  & SR  & 2.082         & 3.224         & 3.633         & 3.653         & 2.878         & 4.020         \\
                  & PR  & 2.739(+0.657)\textuparrow & 2.551(-0.673) & 3.482(-0.151) & 3.739(+0.086)\textuparrow & 3.727(+0.849)\textuparrow & 5.065(+1.045)\textuparrow         \\   \cline{2-8}
 & SIA & 2.245         & 2.367         & 3.469         & 3.857         & 3.694         & 5.367         \\
                  & SFA & 2.510(+0.265)\textuparrow & 2.327(-0.041) & 3.551(+0.082)\textuparrow & 3.776(-0.082) & 3.694(+0.) & 5.286(-0.082)         \\
 \hline
\multirow{4}{*}{GenCN}  & SR  & 2.280         & 1.820         & 4.140         & 4.340         & 2.740         & 3.620         \\
                  & PR  & 2.876(+0.596)\textuparrow & 2.008(+0.188)\textuparrow & 4.200(+0.060)\textuparrow & 4.424(+0.084)\textuparrow & 3.440(+0.700)\textuparrow & 4.464(+0.844)  \\   \cline{2-8}
 & SIA & 2.400         & 1.820         & 4.140         & 4.440         & 3.500         & 4.700         \\
                  & SFA & 2.660(+0.260)\textuparrow & 1.900(+0.080)\textuparrow & 4.280(+0.140)\textuparrow & 4.500(+0.060)\textuparrow & 3.580(+0.080)\textuparrow & 4.600(-0.100)  \\  
\bottomrule
\end{tabular}
}
\label{tab:Narcissistic}
\end{table*}

\begin{figure}[tpb]
    \centering
    \includegraphics[width=0.45\textwidth, height=5.3cm]{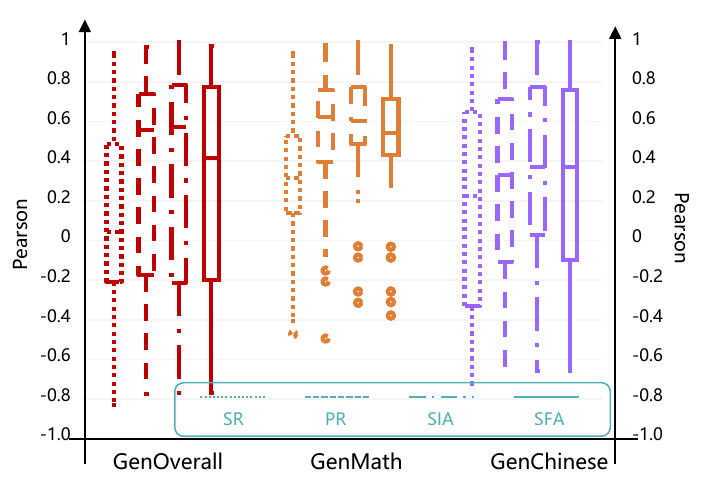}
    \caption{Pearson correlation distributions between model rankings and human preferences under SR, PR, SIA, and SFA. Higher values and narrower distributions indicate stronger and more stable alignment. Additional results are in the Appendix~\ref{appendix:selfbias}.}
    \label{fig:alignment}
    \label{fig:box_corr}
\end{figure}

\begin{figure}[tb]
    \centering
     \includegraphics[width=0.45\textwidth, height=5.3cm]{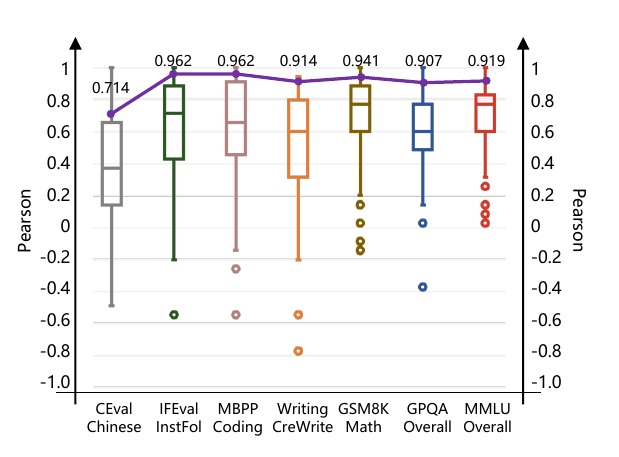}
    \caption{Micro-level Pearson correlation distributions between human rankings and Kemeny-Young aggregated rankings across benchmarks. Purple dots indicate macro-level correlations. “CreWrite” and “InstFol” denote creative writing and instruction following.}

    \label{fig:diff-capable}
\end{figure}
\subsection{Q2: Can Game-Theoretic Re-ranking Mitigate Self-Preference Bias in LLMs?}
\label{sec:eval-protocols}

Recent research has demonstrated that LLMs commonly exhibit \textbf{self-preference bias} when evaluating their own outputs~\cite{DBLP:conf/acl/LiuML24, dietz2025llmevaluationtropesperspectivesvalidity}. In such cases, models tend to assign favorable scores to their own generations, compromising the objectivity of evaluation outcomes. This phenomenon raises concerns about the reliability of LLMs as autonomous evaluators, particularly in decentralized or multi-agent settings where self-assessment can distort collective judgment.

To investigate self preference bias, we compare 4 evaluation protocols that differ in whether self rankings are used and whether rankings are aggregated. In both Self Ranking \textbf{(SR)} and Peer Ranking \textbf{(PR)}, each model produces a complete ranking over all models, and model scores are derived directly from these rankings without applying consensus aggregation. Under \textbf{(SR)}, a model is scored using the rank it assigns to itself, reflecting pure self assessment. Under \textbf{(PR)}, a model is scored by averaging the ranks assigned to it by all peer models, thereby capturing peer based assessment while excluding self judgments. Under Self Inclusive Aggregation \textbf{(SIA)}, each model provides a ranking over all models, including itself. All such rankings are aggregated using the \texttt{Kemeny-Young} method to produce a consensus ranking that reflects collective judgments while retaining the influence of self evaluations. Under Self Free Aggregation \textbf{(SFA)}, self assigned ranks are excluded and treated as missing, and only rankings assigned to other models are retained. These partially observed rankings are aggregated using the same \texttt{Kemeny-Young} procedure, yielding a consensus ranking based solely on peer assessments and isolating the effect of self preference bias.

\paragraph{Game-Theoretic Aggregation Mitigates Self-Preference Bias.}
As shown in Table~\ref{tab:Narcissistic}, rankings produced under SR are consistently more favorable than those obtained from PR across all benchmarks, corroborating observations reported in prior studies on LLM self-evaluation bias~\cite{DBLP:conf/acl/LiuML24, dietz2025llmevaluationtropesperspectivesvalidity}. For example, on GenOverall, \texttt{gpt-4o-20241120} improves by 0.408 ranking positions under peer based evaluation, moving from 2.521 with SR to 2.929 with PR. Similar trends are consistently observed on GenMath and GenChinese. Additional experiments on other benchmarks lead to the same qualitative conclusions and are reported in the Table~\ref{tab:selfbias-otherdata}. To assess whether game theoretic aggregation mitigates this bias, we compare rankings from SIA and SFA using the \texttt{Kemeny-Young} method. In contrast to the pronounced SR–PR gaps, differences between SIA and SFA are consistently small across benchmarks, suggesting that aggregation substantially attenuates the influence of self preference bias and yields more balanced rankings in practice.

To examine alignment with human judgment, Figure~\ref{fig:box_corr} reports the distribution of Pearson correlations between model generated rankings and human preferences. Rankings from the SR show the weakest and most variable alignment, indicating the inherent unreliability of self assigned rankings. In contrast, the SIA achieves the strongest overall alignment, while the SFA yields comparable results, demonstrating that game theoretic aggregation effectively mitigates self preference bias even when self evaluations are included. Due to space limitations, we present results on 3 representative benchmarks in Table~\ref{tab:Narcissistic} and Figure~\ref{fig:box_corr}, with full results provided in Appendix~\ref{appendix:selfbias}.

\subsection{Q3: Which Abilities of LLMs Can Be Reliably Evaluated Through Game Theory?}
\paragraph{Game-Theoretic Evaluation Yields the Most Human-Aligned Rankings in Math Tasks.} Figure~\ref{fig:diff-capable} shows how well game-theoretic aggregation using the Kemeny-Young algorithm aligns with human preferences across different types of LLM capabilities. For each benchmark, we report both micro-level and macro-level Pearson correlations with human rankings from the Chatbot Arena. The box plots display the distribution of micro-level correlations across individual questions, while the purple dots represent macro-level correlations aggregated over the dataset. Among all tasks, the GSM8K benchmark shows the strongest alignment with human judgment. Game-theoretic aggregation achieves both the highest median micro-level correlation and the smallest interquartile range, indicating stable and consistent alignment across questions. At the macro level, it reaches a Pearson correlation of 0.941, outperforming all other datasets. In contrast, benchmarks involving CEval or Creative Writing exhibit lower and more variable correlations, a pattern consistent with the difficulty of capturing human preferences in linguistically nuanced tasks.

These findings suggest that game-theoretic aggregation is particularly effective in tasks with clearer evaluation criteria, such as math and code generation. While it also shows promise in more subjective tasks, additional factors such as task context or stylistic variation may need to be considered to better align with human judgment. Notably, despite high micro-level variance, the \texttt{Creative Writing} benchmark achieves remarkably high macro-level alignment with human preferences (0.914), indicating that aggregate judgments can still reliably reflect human consensus. Overall, the results highlight the applicability of peer-based evaluation across diverse capabilities while also revealing its current limitations.

\section{Conclusion}


We study LLM evaluation from a game-theoretic perspective and argue that assessing model capabilities is fundamentally a ranking aggregation problem rather than a purely score-based comparison. We propose a decentralized peer ranking framework in which LLMs act as both evaluators and evaluatees, and individual rankings are aggregated using principled aggregation methods. Across diverse benchmarks, we show that aggregation-based approaches, particularly consensus-based methods such as Kemeny--Young, achieve stronger and more stable alignment with human judgments than single-model or accuracy-based evaluations, while substantially mitigating self-preference bias. Importantly, these advantages persist under sparse comparison regimes, enabling scalable evaluation without modifying the aggregation procedure. While alignment varies across task types, being strongest for domains with clear correctness signals and weaker for more subjective tasks such as creative writing, aggregate rankings nevertheless capture meaningful human consensus at the dataset level.


\section*{Impact Statement}
This work investigates ranking aggregation methods rooted in social choice theory, which can be viewed as a form of game-theoretic reasoning in a broad sense. These methods aggregate multiple, potentially biased preference rankings into a consensus outcome, analogous to human evaluation scenarios where individual judges may exhibit self-serving or inconsistent behavior. By applying principled aggregation rules, such as Kemeny--Young, the framework mitigates the influence of individual distortions and yields more reliable rankings. The proposed approach contributes to fairer and more robust evaluation of large language models in decentralized and large-scale settings, without introducing new mechanisms for strategic manipulation. Our study focuses on aggregation under fixed preferences and does not address strategic decision-making in the narrower game-theoretic sense, such as incentive design or adaptive strategies, which we leave for future work.

\bibliography{example_paper}
\bibliographystyle{icml2026}

\newpage
\appendix
\onecolumn
\section{Expander-Graph-Based Sparse Evaluation}
\label{append:full}

To further examine the scalability of our evaluation framework, we conduct additional experiments using expander-graph-based sparse comparison structures. Instead of requiring each model to rank the outputs of all others, the expander graph restricts comparisons to a small, well-connected subset per model while preserving global connectivity. This design reduces the evaluation complexity from \( O(M^2) \) to \( O(KM) \), where \( K \ll M \) denotes the average number of comparisons per model.

Experiments are conducted on the \textsc{IFEval}, \textsc{CEval}, \textsc{MBPP}, and \textsc{WritingBench} benchmarks, using 6 language models: \textsc{Qwen2.5-32B-Instruct}, \textsc{Qwen3-32B}, \textsc{DeepSeek-V3.1}, \textsc{DeepSeek-V3.2-Instruct}, \textsc{DeepSeek-V3.2}, and \textsc{DeepSeek-V3-250324}. We consider two comparison settings. In the \textit{full} setting, each model evaluates the outputs of all other models, resulting in a fully connected comparison graph over the 6 models. In the \textit{exp} setting, each model evaluates only three other models, corresponding to an average comparison degree of \( K = 3 \) under the expander-graph-based sparse structure, substantially reducing the comparison budget.

Figure~\ref{fig:expander_scalability} reports micro-level Pearson correlation distributions between human rankings and Kemeny--Young aggregated rankings under both settings. Despite the significantly reduced number of comparisons in the \textit{exp} setting, the resulting rankings remain largely consistent with those obtained from the full comparison graph, indicating that the expander-based design provides a practical trade-off between evaluation cost and ranking fidelity.

\section{Kendall--Tau Distance and Coefficient}

The Kendall--Tau distance between two permutations $\pi$ and $\sigma$ is defined as the number of discordant pairs:
\begin{equation}
D(\pi, \sigma) = \sum_{i < j} \mathbf{1}\left[ \text{sign}(\pi(i) - \pi(j)) \neq \text{sign}(\sigma(i) - \sigma(j)) \right],
\end{equation}
where $\text{sign}(\cdot)$ returns $+1$ or $-1$ for strict orderings.

The Kendall--Tau \textbf{coefficient} measures normalized alignment between two rankings:
\[
\tau(\pi, \sigma) = \frac{C - D}{\binom{n}{2}},
\]
where $C$ and $D$ denote the numbers of concordant and discordant pairs, respectively.

\paragraph{Example.}
For $\pi = [A, B, C]$ and $\sigma = [B, A, C]$, the only discordant pair is $(A, B)$, yielding $D = 1$ and $\tau = \frac{2 - 1}{3} = \frac{1}{3}$.

\section{Kemeny--Young Rank Aggregation}

Given a set of input rankings $\{\pi_k\}_{k=1}^m$, the Kemeny--Young method seeks a consensus ranking $\sigma^*$ that minimizes the total Kendall--Tau distance to all inputs:
\[
\sigma^* = \arg\min_{\sigma} \sum_{k=1}^m D(\sigma, \pi_k).
\]
This optimization problem is NP-hard in general, though effective heuristics are available in practice.

\paragraph{Procedure.}
\begin{enumerate}
    \item \textbf{Pairwise Comparison:} For each pair of candidates $(i, j)$, count how many rankings place $i$ above $j$.
    \item \textbf{Preference Graph Construction:} Build a directed graph where each edge $(i \rightarrow j)$ is weighted by the number of times $i$ is preferred over $j$.
    \item \textbf{Feedback Arc Set Removal:} Remove a minimum-weight set of edges to make the graph acyclic; a topological ordering of the resulting graph yields the consensus ranking.
\end{enumerate}

\paragraph{Example.}
Given rankings $[A, B, C]$, $[B, A, C]$, and $[C, A, B]$, the resulting pairwise preferences induce conflicts among candidates. The ranking that minimizes the total pairwise disagreement is $[B, A, C]$.

In our implementation, we adapt this formulation by maximizing the total Kendall--Tau coefficient $\sum_k \tau(\sigma, \pi_k)$, which is equivalent to minimizing the total Kendall--Tau distance.

\subsection{Theoretical Properties of Kemeny--Young Aggregation}

The strong empirical performance of the Kemeny--Young method can be understood from the structure of its objective, which explicitly optimizes alignment across all pairwise comparisons.

\paragraph{Maximizing Aggregate Pairwise Alignment.}
Minimizing the sum of Kendall--Tau distances is equivalent to minimizing the total number of pairwise inversions between the consensus ranking and the input rankings. Each inversion represents a local disagreement between two candidates. By globally minimizing such disagreements, the Kemeny--Young solution preserves the largest possible fraction of pairwise preferences expressed by the evaluators, making it well suited to settings with noisy or partially inconsistent rankings.

\paragraph{Interpretation as Social Welfare Maximization.}
The Kemeny--Young objective admits a natural social choice interpretation. If each evaluator assigns unit utility to each pairwise ordering it agrees with, then the total Kendall--Tau alignment corresponds to aggregate utility across evaluators. Under this view, the Kemeny--Young ranking maximizes total group utility, providing a principled justification for its robustness in collective ranking scenarios.

\paragraph{Robustness to Conflicting and Biased Rankings.}
Unlike methods based on local elimination or score accumulation, Kemeny--Young optimizes a global objective defined over all pairwise comparisons simultaneously. This global optimization property reduces sensitivity to anomalous or biased individual rankings, including self-preference bias, as no single evaluator can disproportionately influence the outcome. As a result, the method tends to produce stable and human-aligned rankings even when individual inputs are noisy or partially unreliable.

Together, these properties help explain why Kemeny--Young aggregation consistently achieves strong alignment with human judgment in our experiments, particularly in settings characterized by evaluator disagreement and heterogeneous preferences.

\section{Ranking Aggregation Methods}
\label{appendix:rankers}

Table~\ref{tab:rankers} summarizes the ranking aggregation methods evaluated in this work and their core aggregation principles.

\begin{table}[ht]
\centering
\caption{Summary of ranking aggregation methods evaluated in this work.}
\begin{tabular}{p{4cm} p{12cm}}
\toprule
\textbf{Method} & \textbf{Aggregation Principle} \\
\midrule
Average &
Computes the average ranking position of each model across all rankings and orders models by their mean rank. \\

Borda Count~\cite{saari1985optimal} &
Assigns each model a score based on its position in each ranking, with higher-ranked positions receiving more points; models are ranked by total score. \\

Copeland Method~\cite{saari1996copeland} &
Aggregates rankings by counting pairwise wins and losses between models and ranking them by net win count. \\

Dodgson Method~\cite{brandt2009some} &
Measures the minimum number of adjacent swaps required to make a model a Condorcet winner. \\

IRV (Instant-Runoff Voting)~\cite{brandt2009some} &
Iteratively eliminates the model with the fewest first-choice votes and redistributes votes until a complete ranking is obtained. \\

Kendall Aggregation~\cite{kendall1948rank} &
Selects the ranking that maximizes the total Kendall--Tau correlation with the input rankings, operating directly in the ranking space and typically assuming complete rankings. \\

Spearman Aggregation &
Selects the ranking that maximizes the total Spearman correlation with the input rankings. \\

\midrule
Kemeny--Young~\cite{kemeny1959mathematics} &
Finds the ranking that maximizes alignment with pairwise majority preferences induced by the input rankings, equivalently minimizing total pairwise disagreement; this formulation naturally accommodates partial rankings. \\
\bottomrule
\end{tabular}
\label{tab:rankers}
\end{table}

\paragraph{Average Rank Aggregation.}
The average rank method assigns each candidate a score equal to its mean ranking position across evaluators. Formally, the aggregated score of candidate \( i \) is
\[
s_i = \frac{1}{m} \sum_{k=1}^m \pi_k(i),
\]
and candidates are ordered in ascending order of \( s_i \), with lower scores indicating better ranks.

\paragraph{Borda Count.}
Borda count assigns each candidate a score based on its position in each ranking, with higher-ranked positions receiving more points. For \( n \) candidates, the Borda score of candidate \( i \) is
\[
s_i = \sum_{k=1}^m (n - \pi_k(i)),
\]
and candidates are ranked by decreasing \( s_i \).

\paragraph{Copeland Method.}
The Copeland method aggregates rankings via pairwise comparisons. For each pair of candidates \( (i,j) \), let
\[
w(i,j) = \sum_{k=1}^m \mathbf{1}[\pi_k \text{ ranks } i \text{ above } j].
\]
Candidate \( i \) receives one point for each pairwise win \( w(i,j) > w(j,i) \), and the final ranking is determined by the net number of pairwise wins.

\paragraph{Dodgson Method.}
The Dodgson method seeks the ranking that is closest to being a Condorcet winner. For each candidate, it computes the minimum number of adjacent swaps in the input rankings required to make that candidate beat all others in pairwise comparisons. Candidates are ranked by increasing swap distance.
\paragraph{Instant-Runoff Voting (IRV).}
IRV iteratively eliminates the candidate with the fewest first-place votes. At each round, first-choice votes are counted, the candidate with the minimum count is removed, and ballots are redistributed according to the next preferred candidate. The elimination order induces the final ranking.
\paragraph{Kendall--Tau Aggregation.}
Kendall--Tau aggregation selects the ranking that minimizes the total Kendall--Tau distance to the input rankings:
\[
\sigma^* = \arg\min_{\sigma \in \mathcal{S}_n} \sum_{k=1}^m D(\sigma, \pi_k),
\]
where \( D(\cdot,\cdot) \) denotes the Kendall--Tau distance. This formulation operates directly on complete rankings and typically assumes that all input rankings are total orderings.

\paragraph{Spearman Aggregation.}
Spearman aggregation selects the ranking that maximizes total Spearman rank correlation with the input rankings, equivalently minimizing squared rank deviations:
\[
\sigma^* = \arg\min_{\sigma \in \mathcal{S}_n} \sum_{k=1}^m \sum_{i=1}^n \left( \sigma(i) - \pi_k(i) \right)^2.
\]

\paragraph{Kendall--Tau Aggregation vs. Kemeny--Young under Partial Rankings.}
Kendall--Tau aggregation and the Kemeny--Young method are closely related but not identical in general. When all input rankings are complete and strictly ordered, maximizing the total Kendall--Tau correlation is equivalent to the Kemeny--Young objective of minimizing total pairwise disagreement~\cite{kemeny1959mathematics}. However, this equivalence breaks down in the presence of partial or missing rankings, where the two methods differ fundamentally in how they utilize available information.

To illustrate this difference, consider a setting with 6 candidate models and 6 evaluators, whose rankings are given by
\[
\begin{array}{c|cccccc}
 & \text{model A} & \text{model B} & \text{model C} & \text{model D} & \text{model E} & \text{model F} \\
\hline
\pi_1 & 5 & 1 & 2 & 4 & 6 & 3 \\
\pi_2 & 2 & 1 & 3 & 4 & 5 & 6 \\
\pi_3 & 2 & 3 & 1 & 4 & 5 & 6 \\
\pi_4 & 3 & 2 & 1 & 4 & 5 & 6 \\
\pi_5 & 1 & 3 & 2 & 4 & \bot & \bot \\
\pi_6 & 2 & 3 & 1 & 4 & 6 & 5 \\
\end{array}
\]

where \( \bot \) denotes missing ranks. Applying Kendall--Tau aggregation yields the consensus ranking
\(
[3, 2, 1, 4, 5, 6],
\)
whereas the Kemeny--Young method produces
\(
[2, 3, 1, 4, 5, 6].
\)

The discrepancy arises from the fundamentally different treatment of incomplete information. Kendall--Tau aggregation operates on complete rankings and therefore discards any evaluator with missing values, effectively ignoring the partial ranking \( \pi_5 \). In contrast, the Kemeny--Young method aggregates pairwise majority preferences and can incorporate all available pairwise comparisons. Although \( \pi_5 \) does not specify a full ordering, it provides additional pairwise evidence favoring \textit{model A} over \textit{model B}, which alters the majority relation between these two models and leads to a different consensus outcome. This example illustrates that Kendall--Tau aggregation and Kemeny--Young coincide under complete rankings, while Kemeny--Young is more robust to sparse or incomplete evaluations, a property that is particularly important in decentralized peer evaluation settings.

\section{Dataset Statistics and Selection Criteria}
\label{appendix:datasets}

To comprehensively evaluate the reasoning capabilities of LLMs, we adopt a multi-level, multi-dimensional evaluation framework. We carefully select three representative benchmark datasets—GSM8K, MMLU, and GPQA—that span tasks of varying difficulty levels and reasoning demands, as summarized in Table~\ref{tab:dataset_statistics}.

These benchmarks are widely recognized for their effectiveness in probing high-level cognitive capabilities in LLMs, each emphasizing distinct dimensions of reasoning:
\begin{table}[h]
    \centering
    \caption{Key Properties of Evaluation Datasets}
    \label{tab:dataset_statistics}
    \begin{tabular}{l p{4.5cm} c}
        \toprule
        \textbf{Dataset} & \textbf{Domain Focus} & \textbf{Sampled} \\
        \midrule
        CEval & Standardized Exams (Chinese) & 100 \\
        IFEval & Instruction Following & 100 \\
        MBPP & Program Synthesis & 100 \\
        GSM8K & Elementary Math Reasoning & 100 \\
        MMLU & Interdisciplinary Knowledge & 114 \\
        GPQA & Complex Problem Solving & 50 \\
        Creative Writing & Open-ended Generation & 100 \\
        \bottomrule
    \end{tabular}
\end{table}

\begin{itemize}
    \item \textbf{GSM8K}~\cite{DBLP:conf/icml/ChiangZ0ALLZ0JG24} focuses on basic mathematical reasoning. We sample 100 elementary-level arithmetic word problems to assess the model's capabilities in computation, basic logic, and semantic understanding.

    \item \textbf{MMLU}~\cite{DBLP:conf/iclr/HendrycksBBZMSS21} extends the evaluation to interdisciplinary knowledge integration. We select 114 multiple-choice questions across 57 subjects, spanning STEM fields (e.g., mathematics, physics) as well as humanities and social sciences (e.g., history, law), to evaluate the model’s cross-domain reasoning and conceptual abstraction skills.

    \item \textbf{GPQA}~\cite{DBLP:journals/corr/abs-2311-12022} is a high-level benchmark specifically introduced in this study. It comprises 50 open-ended generative questions that require comprehensive problem solving, often involving multimodal information fusion, multi-constraint optimization, and counterfactual reasoning. Models are required to produce full natural language solutions, posing greater demands on deep reasoning, creative thinking, and knowledge transfer.

    \item \textbf{CEval}~\cite{DBLP:conf/nips/HuangBZZZSLLZLF23} is a Chinese-language exam-style benchmark that evaluates model performance on national-level standardized test questions across subjects such as law, medicine, and finance. We use a 100-example subset drawn from its development set, covering both humanities and STEM domains.

    \item \textbf{IFEval}~\cite{DBLP:journals/corr/abs-2311-07911} evaluates instruction-following capabilities of large language models. It includes a diverse range of tasks involving goal specification, constraint satisfaction, and multi-step procedural reasoning. We sample 100 instances across different task categories to assess general instruction comprehension.

    \item \textbf{MBPP}~\cite{DBLP:journals/corr/abs-2108-07732} targets program synthesis. Each instance consists of a short programming task described in natural language along with a set of unit tests. We sample 100 problems to evaluate coding proficiency and functional reasoning ability.

    \item \textbf{Creative Writing}~\cite{DBLP:journals/corr/abs-2503-05244} consists of open-ended generation tasks such as story continuation, character development, and stylistic rewriting. We sample 100 prompts to assess creativity, coherence, and narrative fluency under subjective evaluation.

    \item \textbf{GenMath}. GenMath is a newly constructed generative evaluation set designed to assess mathematical ability. All questions are generated using GPT-4 and are not observed during model pretraining. The prompt templates used for data generation are provided in Section~\ref{appendix:Q4-prompt}.

    \item \textbf{GenChinese}. GenChinese is a generative evaluation set targeting Chinese language understanding and generation. It consists of newly generated questions covering reading comprehension, semantic reasoning, and instruction understanding in Chinese. All instances are generated using GPT-4 and are not observed during model pretraining. The prompt templates used for data generation are provided in Section~\ref{appendix:Q4-prompt}.
    
    \item \textbf{GenOverall}. GenOverall is designed to evaluate general comprehension and reasoning ability across diverse domains. The dataset includes open-ended questions spanning everyday knowledge, abstract reasoning, and cross-domain understanding. All questions are generated using GPT-4 and are not observed during model pretraining. The prompt templates used for data generation are provided in Section~\ref{appendix:Q4-prompt}.
\end{itemize}

\section{Supplementary experiment on Q1}
\label{appendix:appendxq1}
\begin{table*}[ht]
\centering
\caption{Distribution of \textbf{Pearson and Kendall} correlation coefficients between model-generated rankings and human preferences on the \textbf{GSM8K} dataset. Each value reflects \textbf{micro-level alignment}, computed as the correlation between a model-generated ranking and the corresponding human ranking per question. The left columns show the results for individual LLMs, while the right columns report outcomes after applying different aggregation algorithms. Summary statistics include the mean, standard deviation, minimum, quartiles, and maximum. Higher values indicate stronger alignment with human judgment. Color intensity reflects correlation strength. Abbreviations: Avg(average voting), Dod(Dodgson), Cop(Copeland), Bor(Borda), Irv(instant-runoff voting), Spm(Spearman), Kem(Kemeny-Young), Ken(Kendall). Models: 4o11(gpt-4o-20241120), 4o05(gpt-4o-20240513), 4o08(gpt-4o-20240806), Sn10(claude-3.5-Sonnet-20241022), Hk10(claude-3.5-haiku-20241022), Op02(claude-3-opus-20240229).}
\resizebox{\textwidth}{!}{
\begin{tabular}{ll|cccccc|cccccccc}
\toprule
& & \textbf{4o11} & \textbf{Sn10} & \textbf{4o05} & \textbf{4o08} & \textbf{Op02} & \textbf{Hk10} & \textbf{Avg} & \textbf{Dod} & \textbf{Cop} & \textbf{Bor} & \textbf{Irv} & \textbf{Spm} & \textbf{Kem} & \textbf{Ken} \\
\midrule
\multirow{7}{*}{\rotatebox{90}{Pearson}} &\textbf{Mean}  & 0.645 & 0.291 & 0.616 & 0.555 & 0.426 & 0.262 & 0.664 & 0.659 & 0.665 & 0.670 & 0.569 & 0.666 & 0.695 & 0.678 \\
& \textbf{Std.}   & 0.222 & 0.427 & 0.233 & 0.280 & 0.293 & 0.381 & 0.250 & 0.266 & 0.278 & 0.264 & 0.402 & 0.264 & 0.277 & 0.295 \\
& \textbf{Min}   & -0.086 & -0.829 & 0.029 & -0.486 & -0.600 & -0.429 & -0.127 & -0.143 & -0.143 & -0.143 & -0.371 & -0.143 & -0.143 & -0.200 \\
& \textbf{25\%}  & \colorCell{0.486} & \colorCell{0.029} & \colorCell{0.429} & \colorCell{0.371} & \colorCell{0.257} & \colorCell{0.043} & \colorCell{0.562} & \colorCell{0.543} & \colorCell{0.557} & \colorCell{0.600} & \colorCell{0.143} & \colorCell{0.600} & \colorCell{0.600} & \colorCell{0.614} \\
& \textbf{50\%}  & \colorCell{0.714} & \colorCell{0.314} & \colorCell{0.771} & \colorCell{0.657} & \colorCell{0.486} & \colorCell{0.257} & \colorCell{0.764} & \colorCell{0.714} & \colorCell{0.743} & \colorCell{0.771} & \colorCell{0.429} & \colorCell{0.771} & \colorCell{0.771} & \colorCell{0.771} \\
& \textbf{75\%}  & \colorCell{0.771} & \colorCell{0.600} & \colorCell{0.771} & \colorCell{0.771} & \colorCell{0.657} & \colorCell{0.543} & \colorCell{0.809} & \colorCell{0.871} & \colorCell{0.871} & \colorCell{0.829} & \colorCell{0.543} & \colorCell{0.829} & \colorCell{0.886} & \colorCell{0.886} \\
& \textbf{Max}   & 0.943 & 0.943 & 0.886 & 0.943 & 0.829 & 0.943 & 0.975 & 1.000 & 1.000 & 1.000 & 0.734 & 1.000 & 1.000 & 1.000 \\
\hline
\multirow{7}{*}{\rotatebox{90}{Kendall}} &\textbf{Mean} & 0.517 & 0.243   & 0.496 & 0.440 & 0.333   & 0.232  & 0.535   & 0.560 & 0.568 & 0.563  & 0.464  & 0.555 & 0.597  & 0.584   \\
& \textbf{Std.}  & 0.200 & 0.350   & 0.189 & 0.229 & 0.235   & 0.301  & 0.236   & 0.231 & 0.253 & 0.236  & 0.402  & 0.236 & 0.243  & 0.252   \\
& \textbf{Min}  & -0.067  & -0.600  & 0.067 & -0.333  & -0.467  & -0.333 & -0.067  & -0.067   & -0.067   & -0.067 & -0.333 & -0.067   & -0.067 & -0.067  \\
& \textbf{25\%} & \colorCell{0.367} & \colorCell{0.067}   & \colorCell{0.333} & \colorCell{0.333} & \colorCell{0.200}   & \colorCell{0.067}  & \colorCell{0.414}   & \colorCell{0.467} & \colorCell{0.467} & \colorCell{0.467}  & \colorCell{0.067}  & \colorCell{0.467} & \colorCell{0.467}  & \colorCell{0.467}   \\
& \textbf{50\%} & \colorCell{0.600} & \colorCell{0.200}   & \colorCell{0.600} & \colorCell{0.467} & \colorCell{0.333}   & \colorCell{0.200}  & \colorCell{0.574}   & \colorCell{0.600} & \colorCell{0.600} & \colorCell{0.600}  & \colorCell{0.200}  & \colorCell{0.600} & \colorCell{0.600}  & \colorCell{0.600}   \\
& \textbf{75\%} & \colorCell{0.600} & \colorCell{0.467}   & \colorCell{0.600} & \colorCell{0.600} & \colorCell{0.467}   & \colorCell{0.467}  & \colorCell{0.690}   & \colorCell{0.733} & \colorCell{0.733} & \colorCell{0.733}  & \colorCell{0.333}  & \colorCell{0.733} & \colorCell{0.733}  & \colorCell{0.733}   \\
& \textbf{Max}  & 0.867 & 0.867   & 0.733 & 0.867 & 0.733   & 0.867  & 1.000   & 1.000 & 0.467 & 1.000  & 1.000  & 1.000 & 1.000  & 1.000 \\
\bottomrule
\end{tabular}
}

\label{tab:q1-gsm8k}
\end{table*}

\begin{table*}[htbp]
\centering
\caption{
Distribution of \textbf{Pearson} and \textbf{Kendall} correlation coefficients between automated rankings and human ranking on the \textbf{CEval} dataset. Each value reflects micro-level alignment computed per individual problem. The left columns show the results for individual LLMs, while the right columns report outcomes after applying different aggregation algorithms. Summary statistics include the mean, standard deviation, minimum, quartiles, and maximum. Higher values indicate stronger alignment with human judgment. Color intensity reflects correlation strength. Abbreviations: Avg(average voting), Dod(Dodgson), Cop(Copeland), Bor(Borda), Irv(instant-runoff voting), Spm(Spearman), Kem(Kemeny-Young), Ken(Kendall). Models: 4o11(gpt-4o-20241120), 4o05(gpt-4o-20240513), 4o08(gpt-4o-20240806), Sn10(claude-3.5-Sonnet-20241022), Hk10(claude-3.5-haiku-20241022), Op02(claude-3-opus-20240229).}
\resizebox{\textwidth}{!}{
\begin{tabular}{ll|cccccc|cccccccc}
\toprule
& & \textbf{4o11} & \textbf{Sn10} & \textbf{4o05} & \textbf{4o08} & \textbf{Op02} & \textbf{Hk10} & \textbf{Avg} & \textbf{Dod} & \textbf{Cop} & \textbf{Bor} & \textbf{Irv} & \textbf{Spm} & \textbf{Kem} & \textbf{Ken} \\
\midrule
\multirow{7}{*}{\rotatebox{90}{Pearson}} & \textbf{Mean} & 0.376             & 0.165                      & 0.367             & 0.315             & 0.166                  & -0.025                    & 0.317          & 0.351          & 0.326          & 0.321          & -0.083          & 0.301          & 0.365          & 0.335          \\
                  & \textbf{Std.} & 0.348             & 0.334                      & 0.343             & 0.360             & 0.287                  & 0.362                     & 0.396          & 0.385          & 0.377          & 0.367          & 0.468           & 0.372          & 0.396          & 0.385          \\
                  & \textbf{Min}  & -0.371            & -0.486                     & -0.486            & -0.486            & -0.543                 & -0.771                    & -0.576         & -0.486         & -0.486         & -0.486         & -0.829          & -0.486         & -0.486         & -0.486         \\
                  & \textbf{25\%} & \colorCell{0.257}    & -0.086            & \colorCell{0.257}    & \colorCell{0.043}    & \colorCell{0.029}         & -0.329           & \colorCell{0.160} & \colorCell{0.143} & \colorCell{0.143} & \colorCell{0.143} & -0.486 & \colorCell{0.086} & \colorCell{0.143} & \colorCell{0.129} \\
                  & \textbf{50\%} & \colorCell{0.371}    & \colorCell{0.200}             & \colorCell{0.371}    & \colorCell{0.371}    & \colorCell{0.200}         & \colorCell{0.029}            & \colorCell{0.316} & \colorCell{0.371} & \colorCell{0.343} & \colorCell{0.343} & -0.143 & \colorCell{0.286} & \colorCell{0.371} & \colorCell{0.371} \\
                  & \textbf{75\%} & \colorCell{0.600}    & \colorCell{0.357}             & \colorCell{0.643}    & \colorCell{0.600}    & \colorCell{0.371}         & \colorCell{0.257}            & \colorCell{0.597} & \colorCell{0.657} & \colorCell{0.657} & \colorCell{0.600} & \colorCell{0.371}  & \colorCell{0.600} & \colorCell{0.657} & \colorCell{0.600} \\
                  & \textbf{Max}  & 0.943             & 0.829                      & 1.000             & 0.943             & 0.771                  & 0.714                     & 0.968          & 1.000          & 0.943          & 1.000          & 0.886           & 1.000          & 1.000          & 1.000          \\
\hline
\multirow{7}{*}{\rotatebox{90}{Kendall}} & \textbf{Mean} & 0.328             & 0.141                      & 0.307             & 0.275             & 0.142                  & -0.019                    & 0.239          & 0.312          & 0.282          & 0.270          & -0.124          & 0.255          & 0.339          & 0.306          \\
                  & \textbf{Std.} & 0.264             & 0.250                      & 0.275             & 0.297             & 0.202                  & 0.257                     & 0.304          & 0.329          & 0.303          & 0.310          & 0.353           & 0.307          & 0.332          & 0.322          \\
                  & \textbf{Min}  & -0.200            & -0.333                     & -0.333            & -0.333            & -0.333                 & -0.600                    & -0.333         & -0.333         & -0.333         & -0.333         & -0.733          & -0.333         & -0.333         & -0.333         \\
                  & \textbf{25\%} & \colorCell{0.200}    & -0.067            & \colorCell{0.200}    & \colorCell{0.067}    & \colorCell{0.067}         & -0.200           & \colorCell{0.067} & \colorCell{0.067} & \colorCell{0.167} & \colorCell{0.067} & -0.367 & \colorCell{0.067} & \colorCell{0.200} & \colorCell{0.067} \\
                  & \textbf{50\%} & \colorCell{0.333}    & \colorCell{0.067}             & \colorCell{0.333}    & \colorCell{0.333}    & \colorCell{0.133}         & \colorCell{0.000}            & \colorCell{0.200} & \colorCell{0.333} & \colorCell{0.200} & \colorCell{0.267} & -0.133 & \colorCell{0.200} & \colorCell{0.333} & \colorCell{0.333} \\
                  & \textbf{75\%} & \colorCell{0.467}    & \colorCell{0.300}             & \colorCell{0.467}    & \colorCell{0.467}    & \colorCell{0.300}         & \colorCell{0.200}            & \colorCell{0.354} & \colorCell{0.600} & \colorCell{0.467} & \colorCell{0.467} & \colorCell{0.067}  & \colorCell{0.467} & \colorCell{0.600} & \colorCell{0.467} \\
                  & \textbf{Max}  & 0.867             & 0.733                      & 1.000             & 0.867             & 0.600                  & 0.467                     & 0.966          & 1.000          & 0.867          & 1.000          & 0.733           & 1.000          & 1.000          & 1.000         \\
\bottomrule
\end{tabular}
}
\label{tab:rerank-corr-ceval}
\end{table*}

\begin{table*}
\centering
\caption{
Distribution of \textbf{Pearson} and \textbf{Kendall} correlation coefficients between automated rankings and human ranking on the \textbf{GPQA} dataset. Each value reflects micro-level alignment computed per individual problem. The left columns show the results for individual LLMs, while the right columns report outcomes after applying different aggregation algorithms. Summary statistics include the mean, standard deviation, minimum, quartiles, and maximum. Higher values indicate stronger alignment with human judgment. Color intensity reflects correlation strength. Abbreviations: Avg(average voting), Dod(Dodgson), Cop(Copeland), Bor(Borda), Irv(instant-runoff voting), Spm(Spearman), Kem(Kemeny-Young), Ken(Kendall). Models: 4o11(gpt-4o-20241120), 4o05(gpt-4o-20240513), 4o08(gpt-4o-20240806), Sn10(claude-3.5-Sonnet-20241022), Hk10(claude-3.5-haiku-20241022), Op02(claude-3-opus-20240229).}
\resizebox{\textwidth}{!}{
\begin{tabular}{ll|cccccc|cccccccc}
\toprule
& & \textbf{4o11} & \textbf{Sn10} & \textbf{4o05} & \textbf{4o08} & \textbf{Op02} & \textbf{Hk10} & \textbf{Avg} & \textbf{Dod} & \textbf{Cop} & \textbf{Bor} & \textbf{Irv} & \textbf{Spm} & \textbf{Kem} & \textbf{Ken} \\
\midrule
\multirow{7}{*}{\rotatebox{90}{Pearson}} & \textbf{Mean} & 0.466             & 0.419                      & 0.557             & 0.443             & 0.323                  & 0.335                     & 0.562          & 0.551          & 0.556          & 0.559          & 0.110          & 0.544          & 0.592          & 0.569          \\
                  & \textbf{Std.} & 0.355             & 0.378                      & 0.319             & 0.368             & 0.388                  & 0.376                     & 0.304          & 0.318          & 0.288          & 0.299          & 0.392          & 0.323          & 0.285          & 0.310          \\
                  & \textbf{Min}  & -0.371            & -0.543                     & -0.429            & -0.371            & -0.714                 & -0.543                    & -0.328         & -0.543         & -0.371         & -0.143         & -0.829         & -0.371         & -0.371         & -0.371         \\
                  & \textbf{25\%} & \colorCell{0.257}    & \colorCell{0.143}             & \colorCell{0.386}    & \colorCell{0.143}    & \colorCell{0.100}         & \colorCell{0.029}            & \colorCell{0.376} & \colorCell{0.371} & \colorCell{0.371} & \colorCell{0.371} & -0.143         & \colorCell{0.371} & \colorCell{0.486} & \colorCell{0.429} \\
                  & \textbf{50\%} & \colorCell{0.486}    & \colorCell{0.400}             & \colorCell{0.629}    & \colorCell{0.486}    & \colorCell{0.314}         & \colorCell{0.371}            & \colorCell{0.633} & \colorCell{0.600} & \colorCell{0.600} & \colorCell{0.600} & \colorCell{0.143} & \colorCell{0.600} & \colorCell{0.600} & \colorCell{0.600} \\
                  & \textbf{75\%} & \colorCell{0.714}    & \colorCell{0.757}             & \colorCell{0.771}    & \colorCell{0.771}    & \colorCell{0.586}         & \colorCell{0.600}            & \colorCell{0.787} & \colorCell{0.771} & \colorCell{0.771} & \colorCell{0.771} & \colorCell{0.429} & \colorCell{0.771} & \colorCell{0.771} & \colorCell{0.771} \\
                  & \textbf{Max}  & 1.000             & 1.000                      & 1.000             & 1.000             & 1.000                  & 0.943                     & 0.945          & 1.000          & 1.000          & 1.000          & 0.771          & 1.000          & 1.000          & 1.000          \\
\hline
\multirow{7}{*}{\rotatebox{90}{Kendall}} & \textbf{Mean} & 0.371             & 0.347                      & 0.459             & 0.344             & 0.264                  & 0.263                     & 0.420          & 0.450          & 0.448          & 0.448          & -0.045         & 0.437          & 0.497          & 0.472          \\
                  & \textbf{Std.} & 0.330             & 0.343                      & 0.282             & 0.326             & 0.337                  & 0.288                     & 0.290          & 0.296          & 0.282          & 0.284          & 0.318          & 0.297          & 0.279          & 0.292          \\
                  & \textbf{Min}  & -0.333            & -0.333                     & -0.333            & -0.333            & -0.467                 & -0.467                    & -0.276         & -0.467         & -0.333         & -0.200         & -0.733         & -0.200         & -0.333         & -0.333         \\
                  & \textbf{25\%} & \colorCell{0.200}    & \colorCell{0.067}             & \colorCell{0.333}    & \colorCell{0.067}    & \colorCell{0.067}         & \colorCell{0.067}            & \colorCell{0.200} & \colorCell{0.333} & \colorCell{0.333} & \colorCell{0.200} & -0.333         & \colorCell{0.200} & \colorCell{0.333} & \colorCell{0.333} \\
                  & \textbf{50\%} & \colorCell{0.333}    & \colorCell{0.333}             & \colorCell{0.467}    & \colorCell{0.333}    & \colorCell{0.200}         & \colorCell{0.333}            & \colorCell{0.414} & \colorCell{0.467} & \colorCell{0.467} & \colorCell{0.467} & \colorCell{0.067} & \colorCell{0.467} & \colorCell{0.467} & \colorCell{0.467} \\
                  & \textbf{75\%} & \colorCell{0.600}    & \colorCell{0.600}             & \colorCell{0.600}    & \colorCell{0.600}    & \colorCell{0.467}         & \colorCell{0.467}            & \colorCell{0.600} & \colorCell{0.600} & \colorCell{0.600} & \colorCell{0.600} & \colorCell{0.200} & \colorCell{0.600} & \colorCell{0.600} & \colorCell{0.600} \\
                  & \textbf{Max}  & 1.000             & 1.000                      & 1.000             & 1.000             & 1.000                  & 0.867                     & 0.867          & 1.000          & 1.000          & 1.000          & 0.600          & 1.000          & 1.000          & 1.000         \\ 
\bottomrule
\end{tabular}
}
\label{tab:rerank-corr-mmlu}
\end{table*}

\begin{table*}
\centering
\caption{
Distribution of \textbf{Pearson} and \textbf{Kendall} correlation coefficients between automated rankings and human ranking on the \textbf{IFEval} dataset. Each value reflects micro-level alignment computed per individual problem. TThe left columns show the results for individual LLMs, while the right columns report outcomes after applying different aggregation algorithms. Summary statistics include the mean, standard deviation, minimum, quartiles, and maximum. Higher values indicate stronger alignment with human judgment. Color intensity reflects correlation strength. Abbreviations: Avg(average voting), Dod(Dodgson), Cop(Copeland), Bor(Borda), Irv(instant-runoff voting), Spm(Spearman), Kem(Kemeny-Young), Ken(Kendall). Models: 4o11(gpt-4o-20241120), 4o05(gpt-4o-20240513), 4o08(gpt-4o-20240806), Sn10(claude-3.5-Sonnet-20241022), Hk10(claude-3.5-haiku-20241022), Op02(claude-3-opus-20240229).}
\resizebox{\textwidth}{!}{
\begin{tabular}{ll|cccccc|cccccccc}
\toprule
& & \textbf{4o11} & \textbf{Sn10} & \textbf{4o05} & \textbf{4o08} & \textbf{Op02} & \textbf{Hk10} & \textbf{Avg} & \textbf{Dod} & \textbf{Cop} & \textbf{Bor} & \textbf{Irv} & \textbf{Spm} & \textbf{Kem} & \textbf{Ken} \\
\midrule
\multirow{7}{*}{\rotatebox{90}{Pearson}} & \textbf{Mean} & 0.550             & 0.241                      & 0.600             & 0.493             & 0.110                  & -0.048                    & 0.535          & 0.600          & 0.570          & 0.565          & -0.015         & 0.543          & 0.621          & 0.587          \\
                  & \textbf{Std.} & 0.350             & 0.460                      & 0.319             & 0.342             & 0.520                  & 0.440                     & 0.396          & 0.362          & 0.358          & 0.370          & 0.431          & 0.379          & 0.366          & 0.374          \\
                  & \textbf{Min}  & -0.371            & -0.829                     & -0.314            & -0.600            & -0.829                 & -0.943                    & -0.743         & -0.600         & -0.600         & -0.657         & -1.000         & -0.657         & -0.543         & -0.543         \\
                  & \textbf{25\%} & \colorCell{0.386}    & -0.029                     & \colorCell{0.429}    & \colorCell{0.271}    & -0.314                 & -0.371                    & \colorCell{0.338} & \colorCell{0.429} & \colorCell{0.371} & \colorCell{0.371} & -0.314         & \colorCell{0.314} & \colorCell{0.429} & \colorCell{0.371} \\
                  & \textbf{50\%} & \colorCell{0.657}    & \colorCell{0.314}             & \colorCell{0.714}    & \colorCell{0.571}    & \colorCell{0.143}         & -0.029                    & \colorCell{0.638} & \colorCell{0.714} & \colorCell{0.657} & \colorCell{0.600} & \colorCell{0.086} & \colorCell{0.600} & \colorCell{0.714} & \colorCell{0.714} \\
                  & \textbf{75\%} & \colorCell{0.829}    & \colorCell{0.586}             & \colorCell{0.829}    & \colorCell{0.714}    & \colorCell{0.543}         & \colorCell{0.300}            & \colorCell{0.835} & \colorCell{0.829} & \colorCell{0.829} & \colorCell{0.829} & \colorCell{0.257} & \colorCell{0.829} & \colorCell{0.886} & \colorCell{0.829} \\
                  & \textbf{Max}  & 0.943             & 0.943                      & 0.943             & 0.943             & 0.886                  & 0.829                     & 0.959          & 1.000          & 1.000          & 1.000          & 0.771          & 1.000          & 1.000          & 1.000          \\
\hline
\multirow{7}{*}{\rotatebox{90}{Kendall}} & \textbf{Mean} & 0.453             & 0.203                      & 0.501             & 0.397             & 0.094                  & -0.045                    & 0.450          & 0.518          & 0.486          & 0.483          & -0.080         & 0.459          & 0.548          & 0.505          \\
                  & \textbf{Std.} & 0.306             & 0.379                      & 0.289             & 0.297             & 0.412                  & 0.359                     & 0.347          & 0.331          & 0.330          & 0.345          & 0.354          & 0.356          & 0.336          & 0.337          \\
                  & \textbf{Min}  & -0.333            & -0.733                     & -0.200            & -0.467            & -0.733                 & -0.867                    & -0.600         & -0.467         & -0.467         & -0.600         & -1.000         & -0.600         & -0.333         & -0.333         \\
                  & \textbf{25\%} & \colorCell{0.333}    & -0.067            & \colorCell{0.333}    & \colorCell{0.200}    & -0.200                 & -0.333                    & \colorCell{0.215} & \colorCell{0.333} & \colorCell{0.333} & \colorCell{0.333} & -0.333         & \colorCell{0.200} & \colorCell{0.333} & \colorCell{0.333} \\
                  & \textbf{50\%} & \colorCell{0.533}    & \colorCell{0.267}             & \colorCell{0.600}    & \colorCell{0.467}    & \colorCell{0.067}         & -0.067                    & \colorCell{0.467} & \colorCell{0.600} & \colorCell{0.600} & \colorCell{0.467} & -0.067         & \colorCell{0.467} & \colorCell{0.600} & \colorCell{0.600} \\
                  & \textbf{75\%} & \colorCell{0.733}    & \colorCell{0.467}             & \colorCell{0.733}    & \colorCell{0.600}    & \colorCell{0.467}         & \colorCell{0.200}            & \colorCell{0.690} & \colorCell{0.733} & \colorCell{0.733} & \colorCell{0.733} & \colorCell{0.200} & \colorCell{0.733} & \colorCell{0.733} & \colorCell{0.733} \\
                  & \textbf{Max}  & 0.867             & 0.867                      & 0.867             & 0.867             & 0.733                  & 0.733                     & 1.000          & 1.000          & 1.000          & 1.000          & 0.600          & 1.000          & 1.000          & 1.000         \\
\bottomrule
\end{tabular}
}

\label{tab:rerank-corr-ifeval}
\end{table*}

\begin{table*}
\centering
\caption{
Distribution of \textbf{Pearson} and \textbf{Kendall} correlation coefficients between automated rankings and human ranking on the \textbf{MBPP} dataset. Each value reflects micro-level alignment computed per individual problem. The left columns show the results for individual LLMs, while the right columns report outcomes after applying different aggregation algorithms. Summary statistics include the mean, standard deviation, minimum, quartiles, and maximum. Higher values indicate stronger alignment with human judgment. Color intensity reflects correlation strength. Abbreviations: Avg(average voting), Dod(Dodgson), Cop(Copeland), Bor(Borda), Irv(instant-runoff voting), Spm(Spearman), Kem(Kemeny-Young), Ken(Kendall). Models: 4o11(gpt-4o-20241120), 4o05(gpt-4o-20240513), 4o08(gpt-4o-20240806), Sn10(claude-3.5-Sonnet-20241022), Hk10(claude-3.5-haiku-20241022), Op02(claude-3-opus-20240229).
}
\resizebox{\textwidth}{!}{
\begin{tabular}{ll|cccccc|cccccccc}
\toprule
& & \textbf{4o11} & \textbf{Sn10} & \textbf{4o05} & \textbf{4o08} & \textbf{Op02} & \textbf{Hk10} & \textbf{Avg} & \textbf{Dod} & \textbf{Cop} & \textbf{Bor} & \textbf{Irv} & \textbf{Spm} & \textbf{Kem} & \textbf{Ken} \\
\midrule
\multirow{7}{*}{\rotatebox{90}{Pearson}} & \textbf{Mean} & 0.555             & 0.209                      & 0.515             & 0.380             & 0.261                  & -0.016                    & 0.555          & 0.613          & 0.551          & 0.540          & 0.086          & 0.527          & 0.596          & 0.574          \\
                  & \textbf{Std.} & 0.303             & 0.340                      & 0.316             & 0.397             & 0.364                  & 0.439                     & 0.382          & 0.348          & 0.373          & 0.366          & 0.397          & 0.368          & 0.343          & 0.348          \\
                  & \textbf{Min}  & -0.543            & -0.714                     & -0.600            & -0.543            & -0.714                 & -0.771                    & -0.696         & -0.543         & -0.543         & -0.543         & -1.000         & -0.543         & -0.543         & -0.543         \\
                  & \textbf{25\%} & \colorCell{0.429}    & -0.029                     & \colorCell{0.371}    & \colorCell{0.200}    & \colorCell{0.029}         & -0.300                    & \colorCell{0.440} & \colorCell{0.457} & \colorCell{0.371} & \colorCell{0.371} & -0.143         & \colorCell{0.371} & \colorCell{0.457} & \colorCell{0.429} \\
                  & \textbf{50\%} & \colorCell{0.629}    & \colorCell{0.257}             & \colorCell{0.514}    & \colorCell{0.371}    & \colorCell{0.314}         & \colorCell{0.086}            & \colorCell{0.596} & \colorCell{0.657} & \colorCell{0.600} & \colorCell{0.543} & \colorCell{0.200} & \colorCell{0.543} & \colorCell{0.657} & \colorCell{0.600} \\
                  & \textbf{75\%} & \colorCell{0.771}    & \colorCell{0.471}             & \colorCell{0.757}    & \colorCell{0.714}    & \colorCell{0.529}         & \colorCell{0.257}            & \colorCell{0.866} & \colorCell{0.886} & \colorCell{0.857} & \colorCell{0.829} & \colorCell{0.371} & \colorCell{0.829} & \colorCell{0.914} & \colorCell{0.829} \\
                  & \textbf{Max}  & 0.886             & 0.771                      & 0.943             & 0.886             & 0.886                  & 0.886                     & 0.990          & 1.000          & 1.000          & 1.000          & 0.714          & 1.000          & 1.000          & 1.000          \\
\hline
\multirow{7}{*}{\rotatebox{90}{Kendall}} & \textbf{Mean} & 0.447             & 0.157                      & 0.419             & 0.320             & 0.219                  & 0.003                     & 0.436          & 0.549          & 0.489          & 0.464          & 0.004          & 0.450          & 0.538          & 0.518          \\
                  & \textbf{Std.} & 0.246             & 0.273                      & 0.281             & 0.321             & 0.296                  & 0.339                     & 0.326          & 0.301          & 0.326          & 0.316          & 0.301          & 0.309          & 0.312          & 0.312          \\
                  & \textbf{Min}  & -0.333            & -0.600                     & -0.467            & -0.467            & -0.600                 & -0.600                    & -0.414         & -0.333         & -0.333         & -0.333         & -1.000         & -0.333         & -0.333         & -0.333         \\
                  & \textbf{25\%} & \colorCell{0.333}    & -0.067                     & \colorCell{0.233}    & \colorCell{0.200}    & \colorCell{0.067}         & -0.200                    & \colorCell{0.200} & \colorCell{0.400} & \colorCell{0.333} & \colorCell{0.200} & -0.200         & \colorCell{0.200} & \colorCell{0.333} & \colorCell{0.333} \\
                  & \textbf{50\%} & \colorCell{0.467}    & \colorCell{0.200}             & \colorCell{0.467}    & \colorCell{0.333}    & \colorCell{0.200}         & \colorCell{0.067}            & \colorCell{0.414} & \colorCell{0.600} & \colorCell{0.467} & \colorCell{0.467} & \colorCell{0.067} & \colorCell{0.467} & \colorCell{0.600} & \colorCell{0.467} \\
                  & \textbf{75\%} & \colorCell{0.600}    & \colorCell{0.333}             & \colorCell{0.600}    & \colorCell{0.600}    & \colorCell{0.467}         & \colorCell{0.200}            & \colorCell{0.690} & \colorCell{0.800} & \colorCell{0.733} & \colorCell{0.733} & \colorCell{0.200} & \colorCell{0.733} & \colorCell{0.800} & \colorCell{0.733} \\
                  & \textbf{Max}  & 0.733             & 0.600                      & 0.867             & 0.733             & 0.733                  & 0.733                     & 1.000          & 1.000          & 1.000          & 1.000          & 0.467          & 1.000          & 1.000          & 1.000         \\
\bottomrule
\end{tabular}
}
\label{tab:rerank-corr-mbpp}
\end{table*}

\begin{table*}
\centering
\caption{
Distribution of \textbf{Pearson} and \textbf{Kendall} correlation coefficients between automated rankings and human ranking on the \textbf{MMLU} dataset. Each value reflects micro-level alignment computed per individual problem. The left columns show the results for individual LLMs, while the right columns report outcomes after applying different aggregation algorithms. Summary statistics include the mean, standard deviation, minimum, quartiles, and maximum. Higher values indicate stronger alignment with human judgment. Color intensity reflects correlation strength. Abbreviations: Avg(average voting), Dod(Dodgson), Cop(Copeland), Bor(Borda), Irv(instant-runoff voting), Spm(Spearman), Kem(Kemeny-Young), Ken(Kendall). Models: 4o11(gpt-4o-20241120), 4o05(gpt-4o-20240513), 4o08(gpt-4o-20240806), Sn10(claude-3.5-Sonnet-20241022), Hk10(claude-3.5-haiku-20241022), Op02(claude-3-opus-20240229).
}
\resizebox{\textwidth}{!}{
\begin{tabular}{ll|cccccc|cccccccc}
\toprule
& & \textbf{4o11} & \textbf{Sn10} & \textbf{4o05} & \textbf{4o08} & \textbf{Op02} & \textbf{Hk10} & \textbf{Avg} & \textbf{Dod} & \textbf{Cop} & \textbf{Bor} & \textbf{Irv} & \textbf{Spm} & \textbf{Kem} & \textbf{Ken} \\
\midrule
\multirow{7}{*}{\rotatebox{90}{Pearson}} & \textbf{Mean} & 0.656             & 0.389                      & 0.618             & 0.503             & 0.345                  & 0.198                     & 0.648          & 0.681          & 0.664          & 0.662          & 0.050          & 0.644          & 0.707          & 0.675          \\
                  & \textbf{Std.} & 0.267             & 0.359                      & 0.265             & 0.305             & 0.362                  & 0.368                     & 0.243          & 0.237          & 0.252          & 0.249          & 0.406          & 0.258          & 0.236          & 0.253          \\
                  & \textbf{Min}  & -0.200            & -0.829                     & -0.371            & -0.200            & -0.829                 & -0.543                    & -0.138         & -0.429         & -0.200         & -0.086         & -1.000         & -0.257         & 0.029          & -0.086         \\
                  & \textbf{25\%} & \colorCell{0.543}    & \colorCell{0.143}             & \colorCell{0.429}    & \colorCell{0.257}    & \colorCell{0.143}         & -0.029                    & \colorCell{0.557} & \colorCell{0.600} & \colorCell{0.600} & \colorCell{0.543} & -0.143         & \colorCell{0.543} & \colorCell{0.600} & \colorCell{0.600} \\
                  & \textbf{50\%} & \colorCell{0.771}    & \colorCell{0.429}             & \colorCell{0.657}    & \colorCell{0.600}    & \colorCell{0.371}         & \colorCell{0.229}            & \colorCell{0.687} & \colorCell{0.771} & \colorCell{0.771} & \colorCell{0.771} & \colorCell{0.029} & \colorCell{0.714} & \colorCell{0.771} & \colorCell{0.771} \\
                  & \textbf{75\%} & \colorCell{0.829}    & \colorCell{0.643}             & \colorCell{0.829}    & \colorCell{0.771}    & \colorCell{0.600}         & \colorCell{0.471}            & \colorCell{0.824} & \colorCell{0.829} & \colorCell{0.829} & \colorCell{0.829} & \colorCell{0.371} & \colorCell{0.829} & \colorCell{0.829} & \colorCell{0.829} \\
                  & \textbf{Max}  & 1.000             & 1.000                      & 1.000             & 1.000             & 1.000                  & 0.829                     & 0.978          & 1.000          & 1.000          & 1.000          & 0.771          & 1.000          & 1.000          & 1.000          \\
\hline
\multirow{7}{*}{\rotatebox{90}{Kendall}} & \textbf{Mean} & 0.533             & 0.302                      & 0.496             & 0.399             & 0.274                  & 0.164                     & 0.528          & 0.574          & 0.566          & 0.556          & -0.043         & 0.536          & 0.607          & 0.574          \\
                  & \textbf{Std.} & 0.259             & 0.284                      & 0.264             & 0.272             & 0.289                  & 0.280                     & 0.241          & 0.221          & 0.240          & 0.239          & 0.329          & 0.242          & 0.235          & 0.246          \\
                  & \textbf{Min}  & -0.067            & -0.600                     & -0.333            & -0.067            & -0.600                 & -0.467                    & -0.138         & -0.200         & -0.200         & -0.067         & -1.000         & -0.200         & -0.067         & -0.200         \\
                  & \textbf{25\%} & \colorCell{0.333}    & \colorCell{0.067}             & \colorCell{0.333}    & \colorCell{0.200}    & \colorCell{0.067}         & -0.067                    & \colorCell{0.414} & \colorCell{0.467} & \colorCell{0.467} & \colorCell{0.467} & -0.333         & \colorCell{0.467} & \colorCell{0.467} & \colorCell{0.467} \\
                  & \textbf{50\%} & \colorCell{0.600}    & \colorCell{0.333}             & \colorCell{0.533}    & \colorCell{0.467}    & \colorCell{0.200}         & \colorCell{0.200}            & \colorCell{0.552} & \colorCell{0.600} & \colorCell{0.600} & \colorCell{0.600} & -0.067         & \colorCell{0.600} & \colorCell{0.600} & \colorCell{0.600} \\
                  & \textbf{75\%} & \colorCell{0.733}    & \colorCell{0.467}             & \colorCell{0.733}    & \colorCell{0.600}    & \colorCell{0.467}         & \colorCell{0.333}            & \colorCell{0.690} & \colorCell{0.733} & \colorCell{0.733} & \colorCell{0.733} & \colorCell{0.200} & \colorCell{0.733} & \colorCell{0.733} & \colorCell{0.733} \\
                  & \textbf{Max}  & 1.000             & 1.000                      & 1.000             & 1.000             & 1.000                  & 0.733                     & 1.000          & 1.000          & 1.000          & 1.000          & 0.600          & 1.000          & 1.000          & 1.000         \\
\bottomrule
\end{tabular}
}

\label{tab:rerank-corr-mmlu}
\end{table*}

\begin{table*}
\centering
\caption{
Distribution of \textbf{Pearson} and \textbf{Kendall} correlation coefficients between automated rankings and human ranking on the \textbf{Creative Writing} dataset. Each value reflects micro-level alignment computed per individual problem. The left columns show the results for individual LLMs, while the right columns report outcomes after applying different aggregation algorithms. Summary statistics include the mean, standard deviation, minimum, quartiles, and maximum. Higher values indicate stronger alignment with human judgment. Color intensity reflects correlation strength. Abbreviations: Avg(average voting), Dod(Dodgson), Cop(Copeland), Bor(Borda), Irv(instant-runoff voting), Spm(Spearman), Kem(Kemeny-Young), Ken(Kendall). Models: 4o11(gpt-4o-20241120), 4o05(gpt-4o-20240513), 4o08(gpt-4o-20240806), Sn10(claude-3.5-Sonnet-20241022), Hk10(claude-3.5-haiku-20241022), Op02(claude-3-opus-20240229).}
\resizebox{\textwidth}{!}{
\begin{tabular}{ll|cccccc|cccccccc}
\toprule
& & \textbf{4o11} & \textbf{Sn10} & \textbf{4o05} & \textbf{4o08} & \textbf{Op02} & \textbf{Hk10} & \textbf{Avg} & \textbf{Dod} & \textbf{Cop} & \textbf{Bor} & \textbf{Irv} & \textbf{Spm} & \textbf{Kem} & \textbf{Ken} \\
\midrule
\multirow{7}{*}{\rotatebox{90}{Pearson}} & \textbf{Mean} & 0.375             & 0.135                      & 0.465             & 0.303             & 0.197                  & 0.105                     & 0.410          & 0.482          & 0.445          & 0.406          & -0.113         & 0.386          & 0.486          & 0.449          \\
                  & \textbf{Std.} & 0.413             & 0.475                      & 0.328             & 0.469             & 0.377                  & 0.422                     & 0.490          & 0.469          & 0.448          & 0.489          & 0.442          & 0.509          & 0.447          & 0.474          \\
                  & \textbf{Min}  & -0.486            & -0.829                     & -0.543            & -0.600            & -0.600                 & -0.771                    & -0.894         & -0.886         & -0.886         & -0.886         & -1.000         & -0.886         & -0.771         & -0.771         \\
                  & \textbf{25\%} & \colorCell{0.029}    & -0.200                     & \colorCell{0.257}    & -0.071            & -0.086                 & -0.200                    & \colorCell{0.283} & \colorCell{0.371} & \colorCell{0.286} & \colorCell{0.286} & -0.400         & \colorCell{0.257} & \colorCell{0.314} & \colorCell{0.257} \\
                  & \textbf{50\%} & \colorCell{0.371}    & \colorCell{0.086}             & \colorCell{0.429}    & \colorCell{0.429}    & \colorCell{0.114}         & \colorCell{0.114}            & \colorCell{0.643} & \colorCell{0.657} & \colorCell{0.600} & \colorCell{0.543} & -0.086         & \colorCell{0.543} & \colorCell{0.600} & \colorCell{0.600} \\
                  & \textbf{75\%} & \colorCell{0.714}    & \colorCell{0.543}             & \colorCell{0.700}    & \colorCell{0.714}    & \colorCell{0.543}         & \colorCell{0.371}            & \colorCell{0.741} & \colorCell{0.829} & \colorCell{0.743} & \colorCell{0.686} & \colorCell{0.200} & \colorCell{0.686} & \colorCell{0.800} & \colorCell{0.800} \\
                  & \textbf{Max}  & 0.943             & 0.943                      & 0.943             & 1.000             & 0.943                  & 0.886                     & 0.876          & 0.943          & 0.943          & 0.943          & 0.714          & 0.943          & 0.943          & 0.943          \\
\hline
\multirow{7}{*}{\rotatebox{90}{Kendall}} & \textbf{Mean} & 0.299             & 0.105                      & 0.368             & 0.259             & 0.133                  & 0.117                     & 0.296          & 0.406          & 0.368          & 0.338          & -0.153         & 0.312          & 0.424          & 0.381          \\
                  & \textbf{Std.} & 0.347             & 0.385                      & 0.301             & 0.391             & 0.297                  & 0.312                     & 0.433          & 0.401          & 0.386          & 0.412          & 0.357          & 0.426          & 0.385          & 0.415          \\
                  & \textbf{Min}  & -0.333            & -0.733                     & -0.467            & -0.467            & -0.467                 & -0.600                    & -0.828         & -0.733         & -0.733         & -0.733         & -1.000         & -0.733         & -0.600         & -0.600         \\
                  & \textbf{25\%} & \colorCell{0.067}    & -0.200                     & \colorCell{0.200}    & -0.067            & -0.067                 & -0.067                    & \colorCell{0.200} & \colorCell{0.267} & \colorCell{0.200} & \colorCell{0.200} & -0.333         & \colorCell{0.200} & \colorCell{0.267} & \colorCell{0.200} \\
                  & \textbf{50\%} & \colorCell{0.333}    & \colorCell{0.067}             & \colorCell{0.333}    & \colorCell{0.333}    & \colorCell{0.067}         & \colorCell{0.067}            & \colorCell{0.333} & \colorCell{0.467} & \colorCell{0.467} & \colorCell{0.467} & -0.067         & \colorCell{0.333} & \colorCell{0.467} & \colorCell{0.467} \\
                  & \textbf{75\%} & \colorCell{0.467}    & \colorCell{0.333}             & \colorCell{0.567}    & \colorCell{0.600}    & \colorCell{0.333}         & \colorCell{0.333}            & \colorCell{0.600} & \colorCell{0.733} & \colorCell{0.600} & \colorCell{0.600} & \colorCell{0.067} & \colorCell{0.600} & \colorCell{0.667} & \colorCell{0.667} \\
                  & \textbf{Max}  & 0.867             & 0.867                      & 0.867             & 1.000             & 0.867                  & 0.733                     & 0.828          & 0.867          & 0.867          & 0.867          & 0.467          & 0.867          & 0.867          & 0.867         \\
\bottomrule
\end{tabular}
}

\label{tab:rerank-corr-writing}
\end{table*}

\clearpage
\begin{table*}[htbp]
\centering
\caption{\textbf{Distribution of Pearson correlation coefficients} between model-generated rankings and human preferences on the \texttt{GenOverall}, \texttt{GenMath} and \texttt{GenChinese} dataset. Correlations are computed at the micro level as the aglinment between model rankings and human rankings for each question. The left columns show the results for individual LLMs, while the right columns report outcomes after applying different aggregation algorithms. Summary statistics include the mean, standard deviation, minimum, quartiles, and maximum. Higher values indicate stronger alignment with human judgment. Color intensity reflects correlation strength. Abbreviations: Avg(average voting), Dod(Dodgson), Cop(Copeland), Bor(Borda), Irv(instant-runoff voting), Spm(Spearman), Kem(Kemeny-Young), Ken(Kendall). Models: 4o11(gpt-4o-20241120), 4o05(gpt-4o-20240513), 4o08(gpt-4o-20240806), Sn10(claude-3.5-Sonnet-20241022), Hk10(claude-3.5-haiku-20241022), Op02(claude-3-opus-20240229).
}
\resizebox{\textwidth}{!}{
\begin{tabular}{ll|cccccc|ccccccc}
\toprule
& & \textbf{4o11} & \textbf{Sn10} & \textbf{4o05} & \textbf{4o08} & \textbf{Op02} & \textbf{Hk10} & \textbf{Avg} & \textbf{Dod} & \textbf{Cop} & \textbf{Bor} & \textbf{Spm} & \textbf{Kem} & \textbf{Ken} \\
\midrule

\multirow{7}{*}{\rotatebox{90}{\texttt{GenOverall}}} & \textbf{Mean} & 0.307             & 0.163                      & 0.371             & 0.331             & 0.167                  & 0.116                     & 0.384          & 0.452          & 0.401          & 0.414          & 0.390          & 0.436          & 0.425          \\
                  & \textbf{Std.} & 0.449             & 0.446                      & 0.445             & 0.449             & 0.427                  & 0.439                     & 0.453          & 0.389          & 0.418          & 0.432          & 0.446          & 0.426          & 0.431          \\
                  & \textbf{Min}  & -0.771            & -0.771                     & -0.657            & -0.657            & -0.657                 & -0.829                    & -0.683         & -0.657         & -0.657         & -0.657         & -0.657         & -0.657         & -0.657         \\
                  & \textbf{25\%} & -0.029   & -0.143            & -0.029   & -0.029   & -0.143        & -0.200           & \colorCell{0.166} & \colorCell{0.143} & \colorCell{0.143} & \colorCell{0.143} & \colorCell{0.114} & \colorCell{0.114} & \colorCell{0.086} \\
                  & \textbf{50\%} & \colorCell{0.486}    & \colorCell{0.229}             & \colorCell{0.486}    & \colorCell{0.429}    & \colorCell{0.257}         & \colorCell{0.086}            & \colorCell{0.532} & \colorCell{0.600} & \colorCell{0.543} & \colorCell{0.543} & \colorCell{0.486} & \colorCell{0.600} & \colorCell{0.600} \\
                  & \textbf{75\%} & \colorCell{0.643}    & \colorCell{0.543}             & \colorCell{0.714}    & \colorCell{0.686}    & \colorCell{0.543}         & \colorCell{0.457}            & \colorCell{0.736} & \colorCell{0.743} & \colorCell{0.714} & \colorCell{0.743} & \colorCell{0.714} & \colorCell{0.743} & \colorCell{0.743} \\
                  & \textbf{Max}  & 0.943             & 0.829                      & 1.000             & 1.000             & 0.829                  & 0.943                     & 0.983          & 1.000          & 0.943          & 1.000          & 1.000          & 1.000          & 1.000          \\
\hline
\multirow{7}{*}{\rotatebox{90}{\texttt{GenMath}}} & \textbf{Mean} & 0.553             & 0.217                      & 0.506             & 0.239             & 0.203                  & 0.320                     & 0.486          & 0.562          & 0.513          & 0.528          & 0.502          & 0.563          & 0.536          \\
                  & \textbf{Std.} & 0.294             & 0.441                      & 0.324             & 0.415             & 0.426                  & 0.427                     & 0.393          & 0.378          & 0.391          & 0.366          & 0.384          & 0.342          & 0.337          \\
                  & \textbf{Min}  & -0.200            & -0.943                     & -0.429            & -0.714            & -0.886                 & -0.714                    & -0.713         & -0.714         & -0.714         & -0.714         & -0.714         & -0.314         & -0.314         \\
                  & \textbf{25\%} & \colorCell{0.429}    & -0.129            & \colorCell{0.314}    & -0.029   & -0.129        & \colorCell{0.043}            & \colorCell{0.302} & \colorCell{0.429} & \colorCell{0.371} & \colorCell{0.371} & \colorCell{0.257} & \colorCell{0.371} & \colorCell{0.371} \\
                  & \textbf{50\%} & \colorCell{0.600}    & \colorCell{0.257}             & \colorCell{0.543}    & \colorCell{0.200}    & \colorCell{0.257}         & \colorCell{0.371}            & \colorCell{0.559} & \colorCell{0.657} & \colorCell{0.657} & \colorCell{0.600} & \colorCell{0.600} & \colorCell{0.657} & \colorCell{0.600} \\
                  & \textbf{75\%} & \colorCell{0.771}    & \colorCell{0.600}             & \colorCell{0.829}    & \colorCell{0.586}    & \colorCell{0.586}         & \colorCell{0.643}            & \colorCell{0.772} & \colorCell{0.829} & \colorCell{0.771} & \colorCell{0.829} & \colorCell{0.829} & \colorCell{0.829} & \colorCell{0.829} \\
                  & \textbf{Max}  & 0.943             & 0.943                      & 0.943             & 0.943             & 0.943                  & 0.943                     & 0.962          & 1.000          & 1.000          & 1.000          & 0.943          & 1.000          & 1.000          \\
\hline
\multirow{7}{*}{\rotatebox{90}{\texttt{GenChinese}}} & \textbf{Mean} & 0.458             & 0.227                      & 0.431             & 0.425             & 0.274                  & 0.178                     & 0.466          & 0.535          & 0.491          & 0.482          & 0.475          & 0.514          & 0.489          \\
                  & \textbf{Std.} & 0.415             & 0.417                      & 0.412             & 0.428             & 0.400                  & 0.437                     & 0.448          & 0.409          & 0.415          & 0.440          & 0.442          & 0.423          & 0.436          \\
                  & \textbf{Min}  & -0.600            & -0.543                     & -0.771            & -0.771            & -0.771                 & -0.543                    & -0.903         & -0.714         & -0.714         & -0.886         & -0.943         & -0.714         & -0.714         \\
                  & \textbf{25\%} & \colorCell{0.286}    & -0.086            & \colorCell{0.286}    & \colorCell{0.243}    & -0.086        & -0.300           & \colorCell{0.271} & \colorCell{0.386} & \colorCell{0.157} & \colorCell{0.286} & \colorCell{0.286} & \colorCell{0.200} & \colorCell{0.157} \\
                  & \textbf{50\%} & \colorCell{0.571}    & \colorCell{0.371}             & \colorCell{0.543}    & \colorCell{0.543}    & \colorCell{0.429}         & \colorCell{0.257}            & \colorCell{0.563} & \colorCell{0.657} & \colorCell{0.600} & \colorCell{0.600} & \colorCell{0.600} & \colorCell{0.600} & \colorCell{0.600} \\
                  & \textbf{75\%} & \colorCell{0.771}    & \colorCell{0.600}             & \colorCell{0.700}    & \colorCell{0.771}    & \colorCell{0.600}         & \colorCell{0.543}            & \colorCell{0.813} & \colorCell{0.829} & \colorCell{0.829} & \colorCell{0.829} & \colorCell{0.829} & \colorCell{0.829} & \colorCell{0.814} \\
                  & \textbf{Max}  & 1.000             & 0.829                      & 1.000             & 0.943             & 0.886                  & 0.943                     & 0.968          & 1.000          & 1.000          & 1.000          & 1.000          & 1.000          & 1.000         \\
\bottomrule
\end{tabular}
}

\label{tab:q4-math}
\end{table*}

Traditional evaluation of LLMs often depends on \textit{static datasets}, which are fixed sets of benchmark questions accompanied by reference answers, such as MMLU~\cite{DBLP:conf/iclr/HendrycksBBZMSS21} and GSM8K~\cite{DBLP:conf/icml/ChiangZ0ALLZ0JG24}. While these datasets consisten consistent and reproducible comparisons, they suffer from inherent limitations. In particular, many static benchmarks have become publicly available and widely used, increasing the risk of data leakage~\cite{DBLP:journals/corr/abs-2311-01964} and benchmark saturation~\cite{DBLP:journals/corr/abs-2404-18824}, leading to inflated performance estimates.

As shown in Table~\ref{tab:q4-math}, despite being constructed synthetically, the generated datasets yield alignment results highly consistent with those observed in Experiment~\ref{exp:Q1} \textbf{Q1}. Our game-theoretic evaluation framework continues to produce rankings that strongly correlate with human preferences across both Pearson and Kendall metrics. Notably, methods such as Kemeny-Young (\texttt{Kem}) again demonstrate the highest median and upper-quartile correlations, reaffirming their robustness. These findings demonstrate that our peer-based aggregation approach is robust to potential data leakage in the benchmark. Furthermore, it maintains strong alignment with human preferences even on entirely novel, LLM-generated datasets, highlighting its reliability across a wide range of evaluation conditions.

\clearpage
\section{Supplementary experiment on Q2}
\label{appendix:selfbias}

\begin{table*}[htbp]
\centering
\caption{\textbf{Model Rankings under Different Evaluation Protocols.} This table presents model rankings on four representative benchmarks (GSM8K, GPQA, MMLU, CEval, IFEval, MBPP, WritingBench) under four evaluation protocols: \textbf{SR}, \textbf{PR}, \textbf{SIA}, and \textbf{SFA}, where lower values indicate better rankings (1 = best, 6 = worst). The SIA and SFA rankings are computed via Kemeny-Young aggregation, with and without the target model's own votes, respectively. Comparing these values allows us to assess the impact of self-preference and the robustness of aggregation. Models: \textbf{4o-1120} = gpt-4o-20241120, \textbf{4o-0513} = gpt-4o-20240513, \textbf{4o-0806} = gpt-4o-20240806, \textbf{Sn-1022} = Claude-3.5-Sonnet-20241022, \textbf{Hk-1022} = Claude-3.5-Haiku-20241022, \textbf{Op-0229} = Claude-3-Opus-20240229. Due to space constraints, results on additional benchmarks are provided in the appendix.}
\renewcommand\arraystretch{1.2}
\resizebox{\textwidth}{!}{
\begin{tabular}{cccccccc}
\toprule
Dataset                & Method & \textbf{4o-1120} & \textbf{Sn-1022} & \textbf{4o-0513} & \textbf{4o-0806} & \textbf{Hk-1022} & \textbf{Op-0229} \\
\hline
\multirow{4}{*}{GSM8K} & SR     & 2.000             & 3.000& 3.420             & 3.44             & 3.860                     & 5.000                  \\
 & PR     & 2.44(+0.44) \textuparrow     & 2.764(-0.236)              & 3.188(-0.232)     & 3.256(-0.184)     & 4.352(+0.492) \textuparrow             & 5.056(+0.056) \textuparrow          \\ \cline{2-8}
 & SIA    & 1.74             & 2.520& 3.08             & 3.48             & 4.68                     & 5.500                  \\
 & SFA    & 1.920(+0.180) \textuparrow     & 2.700(+0.180) \textuparrow              & 3.060(-0.020)     & 3.320(-0.160)     & 4.720(+0.04) \textuparrow             & 5.460(-0.04)          \\
 \hline
\multirow{4}{*}{GPQA}  & SR     & 1.900             & 3.74& 3.060             & 4.460             & 3.714                     & 4.020                  \\
 & PR     & 1.897(-0.003)     & 3.451(-0.289)              & 2.927(-0.133)     & 4.552(+0.092) \textuparrow     & 3.952(+0.238) \textuparrow             & 4.244(+0.224) \textuparrow         \\   \cline{2-8}
 & SIA    & 1.48             & 3.44& 2.760             & 4.900             & 4.020                     & 4.4                  \\
 & SFA    & 1.520(+0.04) \textuparrow     & 3.520(+0.08) \textuparrow              & 2.820(+0.060) \textuparrow     & 4.78(-0.120) \textuparrow     & 4.08(+0.060) \textuparrow            & 4.24(-0.160)          \\
 \hline
\multirow{4}{*}{MMLU}  & SR     & 1.64             & 3.500& 2.167             & 4.632             & 3.886                     & 4.265                  \\
 & PR     & 2.061(+0.420) \textuparrow     & 3.381(-0.119)              & 2.881(+0.714) \textuparrow     & 4.461(-0.170)     & 3.958(+0.072) \textuparrow             & 4.437(+0.171) \textuparrow          \\   \cline{2-8}
 & SIA    & 1.386             & 3.325& 2.272             & 5.000             & 4.211                     & 4.807                  \\
 & SFA    & 1.649(+0.263) \textuparrow     & 3.377(+0.053) \textuparrow              & 2.535(+0.263) \textuparrow     & 4.702(-0.298)     & 4.211(0.0)             & 4.895(+0.088) \textuparrow \\  
\hline
\multirow{4}{*}{CEval}        & SR     & 3.000             & 3.100& 2.820             & 3.160             & 4.375                     & 3.674                  \\
        & PR     & 3.433(+0.433) \textuparrow     & 2.887(-0.213)              & 3.489(+0.669) \textuparrow    & 2.828(-0.332)     & 4.932(+0.557) \textuparrow            & 3.604(-0.070)          \\ \cline{2-8}
        & SIA    & 3.260             & 2.760& 2.980             & 2.840             & 5.280                     & 3.880                  \\
        & SFA    & 3.500(+0.240) \textuparrow    & 2.620(-0.140)              & 3.220(+0.240) \textuparrow    & 2.680(-0.160)     & 5.120(-0.160)             & 3.740(-0.140)          \\
        \hline
\multirow{4}{*}{IFEval}       & SR     & 1.840             & 3.580& 3.420             & 3.360             & 3.580                     & 3.265                  \\  
        & PR     & 2.756(+0.916) \textuparrow    & 3.160(-0.420)              & 3.423(+0.003) \textuparrow    & 3.512(+0.152) \textuparrow    & 4.843(+1.263) \textuparrow            & 3.692(+0.427) \textuparrow         \\  \cline{2-8}
        & SIA    & 1.760             & 2.880& 3.380             & 3.560             & 5.360                     & 4.060                  \\
        & SFA    & 2.180(+0.420) \textuparrow    & 3.040(+0.160) \textuparrow             & 3.500(+0.120) \textuparrow    & 3.460(-0.100)     & 5.300(-0.060)             & 4.000(-0.060)          \\
        \hline
\multirow{4}{*}{MBPP}         & SR     & 1.729             & 3.320& 3.560             & 3.449             & 3.900                     & 3.460                  \\
        & PR     & 2.557(+0.828) \textuparrow    & 3.528(+0.208) \textuparrow             & 3.273(-0.287)     & 3.590(+0.141) \textuparrow    & 4.486(+0.586) \textuparrow            & 3.872(+0.412) \textuparrow         \\  \cline{2-8}
        & SIA    & 1.580             & 3.120& 3.360             & 3.800             & 4.960                     & 4.180                  \\
        & SFA    & 1.780(+0.200) \textuparrow    & 3.380(+0.260) \textuparrow             & 3.260(-0.100)     & 3.760(-0.040)     & 4.820(-0.140)             & 4.300(+0.120) \textuparrow         \\
\hline
\multirow{4}{*}{Writing} & SR     & 2.420             & 3.314& 3.200             & 3.740             & 3.460                     & 3.825                  \\
        & PR     & 2.665(+0.245) \textuparrow    & 3.215(-0.099)              & 3.482(+0.282) \textuparrow    & 3.552(-0.188)     & 4.385(+0.925) \textuparrow            & 3.952(+0.127) \textuparrow         \\  \cline{2-8}
        & SIA    & 1.960             & 3.040& 3.220             & 3.600             & 4.900                     & 4.280                  \\
        & SFA    & 2.220(+0.260) \textuparrow    & 3.040(0.0)              & 3.300(+0.08) \textuparrow    & 3.480(-0.120)     & 4.880(-0.020)             & 4.280(0.0)         \\
\bottomrule
\end{tabular}
}
\label{tab:selfbias-otherdata}
\end{table*}

\begin{figure}[ht]
    \centering
    \begin{subfigure}{0.48\linewidth}
        \centering
        \includegraphics[width=\linewidth]{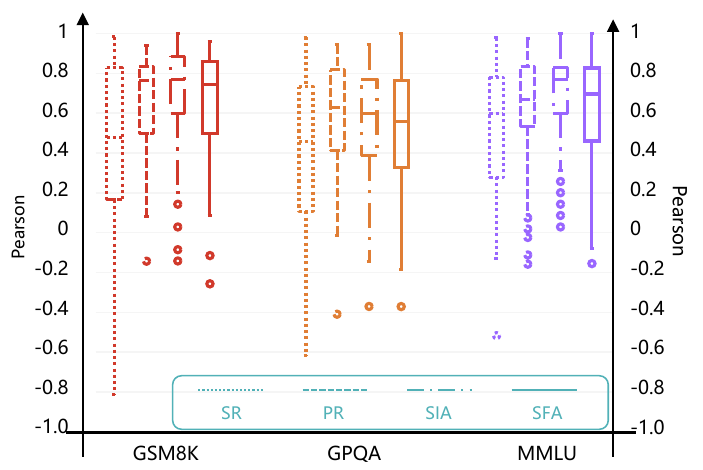}
        \label{fig:genmath}
    \end{subfigure}
    \hfill
    \begin{subfigure}{0.48\linewidth}
        \centering
        \includegraphics[width=\linewidth]{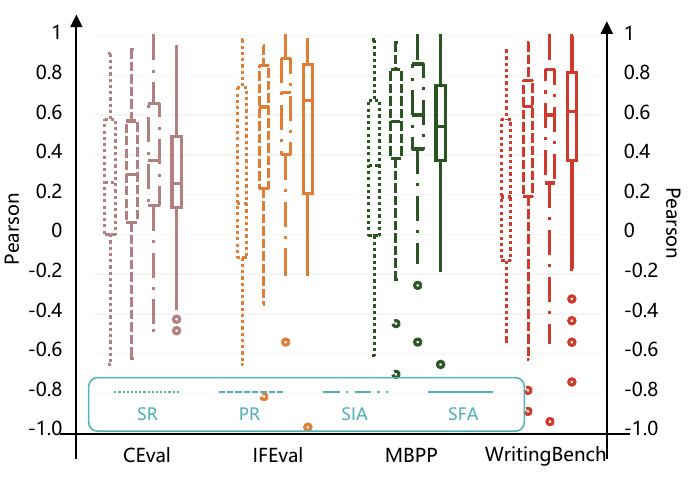}
        \label{fig:genchinese}
    \end{subfigure}
    \caption{\textbf{Alignment with Human Judgments under Different Evaluation Protocols.} This figure reports the distribution of Pearson correlation coefficients between model-generated rankings and human preferences (from Chatbot Arena) across seven benchmarks under four evaluation protocols: \textbf{SR} (Self Ranking), \textbf{PR} (Peer Ranking), \textbf{SIA} (Self-Inclusive Aggregation), and \textbf{SFA} (Self-Free Aggregation). In each boxplot, a \textbf{higher} box indicates stronger alignment with human rankings, while a \textbf{shorter} box implies lower variance and thus more stable alignment. Compared to SR and PR, the SIA and SFA protocols yield both higher and tighter boxes, suggesting that game-theoretic aggregation not only enhances ranking accuracy but also improves robustness against self-bias. Additional results are provided in the appendix due to space constraints.}
    \label{fig:sr_pr}
\end{figure}

\clearpage

\section{Generation Prompts for Capability-Specific Benchmarks}
\label{appendix:Q4-prompt}

\begin{tcolorbox}[
colback=yellow!5!white,
colframe=yellow!75!black,
title=High Difficulty Math Problems Generation Named GenMath,
fonttitle=\bfseries,
boxrule=0.5mm,
arc=2mm,
left=2mm,
right=2mm,
top=2mm,
bottom=2mm,
breakable
]

Please generate \textbf{50 high-difficulty math problems} covering the following fields:

\begin{itemize}
    \item Advanced Algebra
    \item Mathematical Analysis
    \item Number Theory
    \item Combinatorics
    \item Geometry
    \item Differential Equations
    \item Mathematical Logic and Set Theory
    \item Linear Algebra
    \item Probability and Statistics
    \item Elementary Topology
\end{itemize}

\noindent \textbf{Requirements:}

\begin{enumerate}
    \item Output format should be a JSON array in the following structure:
    
    \begin{verbatim}
    [
      {"id": id, "question": question}
    ]
    \end{verbatim}
    
    \item Mathematical expressions and formulas should be written using Markdown math syntax, enclosed in \verb|$...$| for inline math or \verb|$$...$$| for display equations.
    
    \item All questions should be written in \textbf{English}, with clear and precise language.
\end{enumerate}

\end{tcolorbox}

\begin{tcolorbox}[
colback=blue!5!white,
colframe=blue!75!black,
title=High-quality Chinese Question Generation Named GenChinese,
fonttitle=\bfseries,
boxrule=0.5mm,
arc=2mm,
left=2mm,
right=2mm,
top=2mm,
bottom=2mm,
breakable
]

Please generate \textbf{50 Chinese language-related tasks} that cover a comprehensive range of linguistic dimensions. These tasks should be suitable for applications such as:

\begin{itemize}
    \item Phonetics and Phonology 
    \item Vocabulary and Word Formation
    \item Grammar and Syntax 
    \item Rhetoric and Stylistics
    \item Semantics and Pragmatics 
    \item Language Knowledge and Usage
    \item Reading Comprehension
    \item Language Expression and Writing 
\end{itemize}

\noindent \textbf{Requirements:}

\begin{enumerate}
    \item Output format should be a JSON array in the following structure:
    
    \begin{verbatim}
    [
      {"id": id, "question": question}
    ]
    \end{verbatim}
    
    \item All questions should be written in \textbf{Chinese}, with clarity and appropriateness for use in linguistics research, teaching, test design, or LLM training.
\end{enumerate}

\end{tcolorbox}

\begin{tcolorbox}[
colback=teal!5!white,
colframe=teal!75!black,
title=Comprehensive Evaluation of LLM Capabilities Problem Generation Named GenOverall,
fonttitle=\bfseries,
boxrule=0.5mm,
arc=2mm,
left=2mm,
right=2mm,
top=2mm,
bottom=2mm,
breakable
]

Please generate \textbf{50 evaluation questions} designed to comprehensively assess the capabilities of large language models (LLMs). The questions should span a wide range of skills and reasoning dimensions.

\noindent \textbf{Requirements:}

\begin{enumerate}
    \item Output format should be a JSON array in the following structure:
    
    \begin{verbatim}
    [
      {"id": id, "question": question}
    ]
    \end{verbatim}
    
    \item All questions should be written in \textbf{English}, using clear, precise, and instruction-oriented language.
\end{enumerate}

\end{tcolorbox}

\clearpage
\section{Prompts}
\label{appendix:prompts}

\begin{tcolorbox}[
colback=purple!5!white,
colframe=purple!75!black,
title=Answer Ranking Prompt Design for Overall,
fonttitle=\bfseries,
boxrule=0.5mm,
arc=2mm,
left=2mm,
right=2mm,
top=2mm,
bottom=2mm,
breakable
]

You are a reviewer assigned to rank multiple solutions to a given question. Your evaluation must be based solely on the following three criteria:
\begin{itemize}
    \item \textbf{Accuracy}: How correct and relevant is the information?
    \item \textbf{Logical Consistency}: How coherent and well-reasoned is the explanation?
    \item \textbf{Fluency}: How clear and natural is the language?
\end{itemize}

Please strictly follow the format below:

Here is the question and the options:

[Question]

\{\{question.strip()\}\}

\{\% for choice in choices \%\}

\{\{choice\}\}

\{\% endfor \%\}

---

[Solution - 1]

\{\{resps[0]\}\}

[Solution - 2]

\{\{resps[1]\}\}

[Solution - 3]

\{\{resps[2]\}\}

[Solution - 4]

\{\{resps[3]\}\}

[Solution - 5]

\{\{resps[4]\}\}

[Solution - 6]

\{\{resps[5]\}\}

\textbf{Output Format (Rank from best to worst):}

1. Solution x

2. Solution y

3. Solution z

...

You must rank \textbf{all six solutions}, without skipping or tying any of them. \textbf{Do not add any comments or explanations.} Only return the final ordered list by solution number.

\end{tcolorbox}

\begin{tcolorbox}[
colback=yellow!5!white,
colframe=yellow!75!black,
title=Answer Ranking Prompt Design for Mathematical Problem,
fonttitle=\bfseries,
boxrule=0.5mm,
arc=2mm,
left=2mm,
right=2mm,
top=2mm,
bottom=2mm,
breakable
]

You are a reviewer assigned to rank multiple solutions to the same math problem. Your evaluation must be based solely on the following three criteria:
\begin{itemize}
    \item \textbf{Accuracy}: Is the mathematical reasoning correct, and does the solution produce the correct answer?
    \item \textbf{Logical Rigor}: Is the problem-solving process well-structured, justified, and logically sound at each step?
    \item \textbf{Clarity of Explanation}: Is the reasoning clearly explained, using appropriate notation and terminology?
\end{itemize}

Please strictly follow the format below:

Here is the question and the options:

[Question]

\{\{question.strip()\}\}

\{\% for choice in choices \%\}

\{\{choice\}\}

\{\% endfor \%\}

---

[Solution - 1]

\{\{resps[0]\}\}

[Solution - 2]

\{\{resps[1]\}\}

[Solution - 3]

\{\{resps[2]\}\}

[Solution - 4]

\{\{resps[3]\}\}

[Solution - 5]

\{\{resps[4]\}\}

[Solution - 6]

\{\{resps[5]\}\}

\textbf{Output Format (Rank from best to worst):}

1. Solution x

2. Solution y

3. Solution z

...

You must rank \textbf{all six solutions}, without skipping or tying any of them. \textbf{Do not add any comments or explanations.} Only return the final ordered list by solution number.

\end{tcolorbox}

\begin{tcolorbox}[
colback=green!5!white,
colframe=green!75!black,
title=Answer Ranking Prompt Design for Chinese,
fonttitle=\bfseries,
boxrule=0.5mm,
arc=2mm,
left=2mm,
right=2mm,
top=2mm,
bottom=2mm,
breakable
]

You are a reviewer assigned to rank multiple answers written in Chinese. Your evaluation must be based solely on the following three criteria:
\begin{itemize}
    \item \textbf{Linguistic Accuracy}: Are the grammar, vocabulary, and expressions consistent with standard modern Chinese?
    \item \textbf{Clarity of Expression}: Is the language smooth, natural, and easy to understand? Is the logic clearly conveyed?
    \item \textbf{Contextual Appropriateness}: Does the response match the intended tone, audience, and context of the prompt?
\end{itemize}

Please strictly follow the format below:

Here is the question and the options:

[Question]

\{\{question.strip()\}\}

\{\% for choice in choices \%\}

\{\{choice\}\}

\{\% endfor \%\}

---

[Solution - 1]

\{\{resps[0]\}\}

[Solution - 2]

\{\{resps[1]\}\}

[Solution - 3]

\{\{resps[2]\}\}

[Solution - 4]

\{\{resps[3]\}\}

[Solution - 5]

\{\{resps[4]\}\}

[Solution - 6]

\{\{resps[5]\}\}

\textbf{Output Format (Rank from best to worst):}

1. Solution x

2. Solution y

3. Solution z

...

You must rank \textbf{all six solutions}, without skipping or tying any of them. \textbf{Do not add any comments or explanations.} Only return the final ordered list by solution number.

\end{tcolorbox}

\begin{tcolorbox}[
colback=red!5!white,
colframe=red!75!black,
title=Answer Ranking Prompt Design for Instruction Following,
fonttitle=\bfseries,
boxrule=0.5mm,
arc=2mm,
left=2mm,
right=2mm,
top=2mm,
bottom=2mm,
breakable
]

You are a reviewer assigned to rank multiple responses to the same instruction. Your evaluation must be based solely on the following three criteria:
\begin{itemize}
    \item \textbf{Task Completion}: Does the response fully and accurately follow all aspects of the given instruction?
    \item \textbf{Interpretation Accuracy}: Does the response show a correct understanding of the instruction's intent?
    \item \textbf{Relevance and Focus}: Is the content tightly aligned with the instruction, without going off-topic or omitting key parts?
\end{itemize}

Please strictly follow the format below:

Here is the question and the options:

[Question]

\{\{question.strip()\}\}

\{\% for choice in choices \%\}

\{\{choice\}\}

\{\% endfor \%\}

---

[Solution - 1]

\{\{resps[0]\}\}

[Solution - 2]

\{\{resps[1]\}\}

[Solution - 3]

\{\{resps[2]\}\}

[Solution - 4]

\{\{resps[3]\}\}

[Solution - 5]

\{\{resps[4]\}\}

[Solution - 6]

\{\{resps[5]\}\}

\textbf{Output Format (Rank from best to worst):}

1. Solution x

2. Solution y

3. Solution z

...

You must rank \textbf{all six solutions}, without skipping or tying any of them. \textbf{Do not add any comments or explanations.} Only return the final ordered list by solution number.

\end{tcolorbox}

\begin{tcolorbox}[
colback=blue!5!white,
colframe=blue!75!black,
title=Answer Ranking Prompt Design for Code Implementation,
fonttitle=\bfseries,
boxrule=0.5mm,
arc=2mm,
left=2mm,
right=2mm,
top=2mm,
bottom=2mm,
breakable
]

You are a reviewer assigned to rank multiple code implementations. Your evaluation must be based solely on the following three criteria:
\begin{itemize}
    \item \textbf{Correctness}: Does the code run successfully and meet all functional requirements of the prompt?
    \item \textbf{Logical Clarity}: Is the code logically structured and easy to follow?
    \item \textbf{Readability}: Are variable names meaningful, comments helpful, and formatting clean and maintainable?
\end{itemize}

Please strictly follow the format below:

Here is the question and the options:

[Question]

\{\{question.strip()\}\}

\{\% for choice in choices \%\}

\{\{choice\}\}

\{\% endfor \%\}

---

[Solution - 1]

\{\{resps[0]\}\}

[Solution - 2]

\{\{resps[1]\}\}

[Solution - 3]

\{\{resps[2]\}\}

[Solution - 4]

\{\{resps[3]\}\}

[Solution - 5]

\{\{resps[4]\}\}

[Solution - 6]

\{\{resps[5]\}\}

\textbf{Output Format (Rank from best to worst):}

1. Solution x

2. Solution y

3. Solution z

...

You must rank \textbf{all six solutions}, without skipping or tying any of them. \textbf{Do not add any comments or explanations.} Only return the final ordered list by solution number.

\end{tcolorbox}

\begin{tcolorbox}[
colback=purple!5!white,
colframe=purple!75!black,
title=Answer Ranking Prompt Design for Creative Writing,
fonttitle=\bfseries,
boxrule=0.5mm,
arc=2mm,
left=2mm,
right=2mm,
top=2mm,
bottom=2mm,
breakable
]

You are a reviewer assigned to rank multiple creative writing pieces. Your evaluation must be based solely on the following three criteria:
\begin{itemize}
    \item \textbf{Originality}: Is the content imaginative and unique? Does it offer a fresh perspective or concept?
    \item \textbf{Structural Coherence}: Is the narrative or composition logically organized and well-developed from beginning to end?
    \item \textbf{Expressive Quality}: Is the language vivid, engaging, and emotionally resonant?
\end{itemize}

Please strictly follow the format below:

Here is the question and the options:

[Question]

\{\{question.strip()\}\}

\{\% for choice in choices \%\}

\{\{choice\}\}

\{\% endfor \%\}

---

[Solution - 1]

\{\{resps[0]\}\}

[Solution - 2]

\{\{resps[1]\}\}

[Solution - 3]

\{\{resps[2]\}\}

[Solution - 4]

\{\{resps[3]\}\}

[Solution - 5]

\{\{resps[4]\}\}

[Solution - 6]

\{\{resps[5]\}\}

\textbf{Output Format (Rank from best to worst):}

1. Solution x

2. Solution y

3. Solution z

...

You must rank \textbf{all six solutions}, without skipping or tying any of them. \textbf{Do not add any comments or explanations.} Only return the final ordered list by solution number.

\end{tcolorbox}

\end{document}